\documentclass[sigconf]{acmart}

\AtBeginDocument{%
	\providecommand\BibTeX{{%
			\normalfont B\kern-0.5em{\scshape i\kern-0.25em b}\kern-0.8em\TeX}}}

\copyrightyear{2019}
\acmYear{2019}
\setcopyright{acmcopyright}

\copyrightyear{2019} 
\acmYear{2019} 
\acmConference[DAI '19]{First International Conference on Distributed Artificial Intelligence}{October 13--15, 2019}{Beijing, China}
\acmBooktitle{First International Conference on Distributed Artificial Intelligence (DAI '19), October 13--15, 2019, Beijing, China}
\acmPrice{15.00}
\acmDOI{10.1145/3356464.3357706}
\acmISBN{978-1-4503-7656-3/19/10}

\usepackage[utf8]{inputenc}
\usepackage{hyperref}
\usepackage{url}
\usepackage{booktabs}
\usepackage{amsfonts}
\usepackage{amsmath}
\usepackage{microtype}
\usepackage{algorithm}
\usepackage{algorithmic}

\usepackage{subfigure}

\newtheorem{lemma}{Lemma}
\newcommand{\fig}[1]{Fig.~\ref{#1}}

\newcommand{\tb}[1]{Tab.~\ref{#1}}
\newcommand{\se}[1]{Section~\ref{#1}}

\newcommand{\lm}[1]{Lemma~\ref{#1}}
\newcommand{\alg}[1]{Algo.~\ref{#1}}
\newcommand{\ap}[1]{Appendix~\ref{#1}}

\begin{document}
	
	\title{Generative Adversarial Exploration for Reinforcement Learning}
	
	
	\author{Weijun Hong}
	\affiliation{%
		\institution{Shanghai Jiao Tong University}
		\streetaddress{800 Dongchuan Rd.}
	}
	\email{wiljohn@apex.sjtu.edu.cn}
	
	\author{Menghui Zhu}
	\affiliation{%
		\institution{Shanghai Jiao Tong University}
		\streetaddress{800 Dongchuan Rd.}
	}
	\email{zerozmi7@sjtu.edu.cn}
	
	\author{Minghuan Liu}
	\affiliation{%
		\institution{Shanghai Jiao Tong University}
		\streetaddress{800 Dongchuan Rd.}
	}
	\email{minghuanliu@apex.sjtu.edu.cn}
	
	\author{Weinan Zhang}
	\affiliation{%
		\institution{Shanghai Jiao Tong University}
		\streetaddress{800 Dongchuan Rd.}
	}
	\email{wnzhang@sjtu.edu.cn}
	
	\author{Ming Zhou}
	\affiliation{%
		\institution{Shanghai Jiao Tong University}
		\streetaddress{800 Dongchuan Rd.}
	}
	\email{mingak@sjtu.edu.cn}
	
	\author{Yong Yu}
	\affiliation{%
		\institution{Shanghai Jiao Tong University}
		\streetaddress{800 Dongchuan Rd.}
	}
	\email{yyu@apex.sjtu.edu.cn}
	
	\author{Peng Sun}
	\affiliation{%
		\institution{Tencent AI Lab}
		\streetaddress{800 Dongchuan Rd.}
	}
	\email{pengsun000@gmail.com}
	
	\renewcommand{\shortauthors}{Hong, et al.}
	
	\begin{abstract}
		Exploration is crucial for training the optimal reinforcement learning (RL) policy, where the key is to discriminate whether a state visiting is novel. Most previous work focuses on designing heuristic rules or distance metrics to check whether a state is novel without considering such a discrimination process that can be learned.
		In this paper, we propose a novel method called generative adversarial exploration (GAEX) to encourage exploration in RL via introducing an intrinsic reward output from a generative adversarial network, where the generator provides fake samples of states that help discriminator identify those less frequently visited states. Thus the agent is encouraged to visit those states which the discriminator is less confident to judge as visited.
		GAEX is easy to implement and of high training efficiency. 
		In our experiments, we apply GAEX into DQN and the DQN-GAEX algorithm achieves convincing performance on challenging exploration problems, including the game \textit{Venture}, \textit{Montezuma's Revenge} and \textit{Super Mario Bros}, without further fine-tuning on complicate learning algorithms. To our knowledge, this is the first work to employ GAN in RL exploration problems.
	\end{abstract}


	\maketitle
	
	\section{Introduction}
	Reinforcement learning (RL) enables the agent to learn the optimal policy in a trial-and-error manner interacting with the environment \cite{sutton2018reinforcement}, and exploration, which is concerned about making the policy visit diverse states in RL, is crucial for the optimal policy training \cite{strehl2008analysis}.
	However, many existing RL methods only deploy very simple exploration strategies, such as $\epsilon$-greedy in deep Q-network (DQN) \cite{mnih2015human} and the addition of Gaussian noise on policy in deep deterministic policy gradient (DDPG) \cite{lillicrap2015continuous}. Such naive exploration strategies succeed only when the reward is dense (or well-shaped) and the state transitions are simple, and in many hard exploration environment with sparse reward signal, they tend to fail since it is highly difficult to effectively update the agent's policy.
	
	To encourage exploration, normally one needs to design a well-shaped reward function for the policy training.
	However, it is notoriously challenging to find such an extrinsic reward (i.e., reward directly from the environment) function which leads to optimal solutions for various RL tasks. 
	Hence, most research work focuses on supplementing by intrinsic reward \cite{oudeyer2009intrinsic,schmidhuber2010formal}, which is also called exploration bonus, as an alternative to reward engineering, mostly inspired by the concepts of curiosity and surprise \cite{schmidhuber1991curious,itti2006bayesian}. 
	It is just like when human play games, we not only concentrate on maximizing the accumulative rewards, but also keep a rough impression about the visited game scenarios, and feel curious when unseen situations are encountered. Research work inspired by such idea includes methods based on counts or pseudo-counts \cite{kolter2009near,bellemare2016unifying,machado2018count}, information theory \cite{mohamed2015variational,houthooft2016vime} and prediction error \cite{pathak2017curiosity,burda2018exploration}. 

	Essentially, intrinsic-motivated exploration aims to build a state distribution $\rho_\pi(s)$ under policy $\pi$ and determine how novel a given state is, i.e. $P_\pi(\mathbf{1}(s\mbox{ is novel})|s)$, through $\rho_\pi(s)$.
	However, most methods mentioned above model the novelty in a separate manner. 
	Count-based methods, for example, first take state (or state-action pairs) visit counts as visitation frequency of states, and then derive the novelty through it. 
	Information-theoretic methods model the distribution in a Bayesian way by first updating a dynamic model of the environment, and then measuring the distance (e.g. the Kullback-Leibler divergence) when a new state is encountered.
	Prediction-error based methods usually first employ a $\theta$-parametrized predicting model to predict the future state $s_{t+1}$ given the action $a_t$ and state $s_t$ at current time $t$, i.e., $P_\theta(s_{t+1} | s_t, a_t)$, and the curiosity is then given by the deviation from correctly predicting the $s_{t+1}$. 
	However, in order to measure the novelty of an encountered state $s$, it is much more straightforward and reasonable to directly get the probability $P(1(s\mbox{ is novel})|s)$ or $P(1(s\mbox{ is visited})|s)$ which can be provided by a discriminative model. Motivated by such an idea, we introduce exploration bonuses in a more direct way by using generative adversarial network (GAN) \cite{goodfellow2014generative} to provide a credible judgment on state novelty.
	
	GAN is one of the most popular deep learning models in the past few years,
	where a generator $G$ and a discriminator $D$ pit against each other. During the training, $G$ learns to fool $D$ by generating high quality data, while $D$ evolves to distinguish the fake from the real. 
	Finally, both $D$ and $G$ converge when data generated from $G$ follows the similar distribution of the real data and $D$ cannot judge whether the data is fake.
	Because GAN is able to fit the real data distribution, it is promising to generalize this ability into RL exploration problems, i.e. to model the state distribution under the current policy and determine whether a given state is likely to be novel or visited.
	
	In this paper, we propose to use Generative Adversarial Exploration (GAEX) framework that applies GAN to provide the curiosity for exploration, where the discriminator $D$ is trained with real states sampled from the environment, and the fake states are generated from the generative model $G$. The probability of being a real visited state is directly given by $D$ , and is used as an exploration bonus.
	In this setting, we hope those frequently visited states would be marked high probabilities to be real by $D$, which means small exploration bonuses should be related to such states, and on the contrary, a rarely encountered state would be assigned with a low probability as real, which means this state should be paid more attention to explore.
	In addition, our GAEX framework requires no domain knowledge or preassumption, which is required in most previous work including count-based methods and information-theoretic methods. Leveraging the representation learning capability of deep neural network, raw pixels can be directly fed into the network without any feature engineering. Moreover, GAEX is an efficient algorithm for its low computational cost therefore can be easily generalized to expand the existing learning algorithms.
	
	Our experimental results in various environments demonstrate that GAEX is effective and efficient. In a simple chain Markov decision process (MDP) environment, we verify the effectiveness and the stability of GAEX. We also evaluate GAEX in many complex hard-exploration environments including classic Atari games like \textit{Venture} and \textit{Montezuma's Revenge}, and famous Nintendo game \textit{Super Mario Bros} (without extrinsic rewards), where we observe GAEX makes active exploration, and even the state-of-the-art performance on the game \textit{Venture}.
	
	\section{Related Work}
	
	The exploration and exploitation dilemma, especially in sparse reward environments, remains to be a non-trivial problem \cite{ishii2002control}. Practical RL algorithms often explore with simple heuristics, such as $\epsilon$-greedy, Boltzmann exploration and random noise, which give rise to inefficient exploration like random walk. Intrinsic motivation reward is a general solution aiming to provide qualitative guidance to explore states that bring more surprise or reduce the uncertainty \cite{oudeyer2009intrinsic, schmidhuber2010formal}.
	
	Extrinsic rewards can be regarded as the reward signals received from the environment, while the intrinsic rewards are produced by the agent itself. A big group of approaches use state (or state-action pair) visitation count to give intrinsic bonus, encourage to explore less visited states. Classic count-based methods \cite{strehl2005theoretical,strehl2008analysis,kolter2009near} record the state-action occurrence times and solve an approximate Bellman equation before the agent takes an action. These counts, however, are obviously not suitable for large discrete or continuous state/action spaces. To address this issue, \cite{tang2017exploration} introduces a hash function combined with an auto-encoder to reduce the dimension of state space. 
	Further, pseudo-count with a density model is proposed  \cite{bellemare2016unifying,ostrovski2017count}, which requires efforts to hold a statistic model of the distribution of states. 
	Besides these count-based methods, information theory is widely used to denote the uncertainty reduction of visiting a state, such as maximizing information gain \cite{houthooft2016vime}, empowerment (or called mutual information) \cite{mohamed2015variational,still2012information}, etc. These information-theoretic approaches are based on a dynamic model of the environment which predicts the next state $s_{t+1}$ given the current history $\xi_t={s_1,a_1,...,s_t}$ and action $a_t$. However, both state distribution model and environmental dynamics model are hard to build in high-dimensional continuous control task and also limited in specific situations.
	
	Recently, some prediction-error based methods make curiosity-driven exploration popular in a period of time. \cite{pathak2017curiosity} learns an inverse prediction model and takes the difference between predicted state feature $\hat{\phi}(s_{t+1})$
	and real state feature $\phi(s_{t+1})$ to represent the curiosity. \cite{burda2018exploration} also uses the prediction error between a trained model and a fixed randomly initialized neural network. These methods use heuristically designed functions to represent the novelty, which treat a state estimation as an expectation and take the gap between the expectation and the reality or fixed prediction as a guidance to encourage the exploration.
	
	In essence, all the above approaches try to fit the state distribution given certain policy which is used to calculate the curiosity. We consider that using a discriminative model to judge the novelty of state is a more straightforward way to guide the exploration. This requires a discriminator fed with sampled data from experience, and also necessarily negative samples that can be produced by a generator, which comes to our proposed GAEX framework.

	\section{Generative Adversarial Exploration}
	
	\subsection{Background}
	
	In a general RL framework, there is an agent learning in an environment modeled by a Markov Decision Process (MDP) $\mathcal{M} = \{ \mathcal{S}, \mathcal{A}, \mathcal{R}, \mathcal{P}, \gamma \}$, in which $\mathcal{S}$ is the state space, $\mathcal{A}$ the action space, $\mathcal{R}$ the reward function, $\mathcal{P}$ the transition probability distribution and $\gamma$ the future reward discounted factor. A learning agent observes state $s_t$ at timestep $t$, then interacts with the environment by taking action $a_t$, receives an extrinsic reward $r^e_t$, and transitions to a new state $s_{t+1}$. For a finite-horizontal MDP, the goal of a general RL agent is to find an optimal policy $\pi^*$ that maximizes the total expected discounted reward within the horizon $T$ as
	\vspace{-1pt}
	\begin{align}
	\pi^* = \arg\max_{\pi}\mathbb{E}_{\pi,\mathcal{P}}\left[ \sum_{t=0}^T\gamma^t r^e_t\right].
	\end{align}
	When the extrinsic reward $r^e_t$ is sparse, a bonus intrinsic reward $r^i_t$, which is usually related to the novelty of a state, 
	is used at the same time to encourage exploration, then the agent's learning objective change to maximize the augmented total expected discounted reward
	\begin{align}
	\pi^* = \arg\max_{\pi}\mathbb{E}_{\pi,\mathcal{P}}\left[ \sum_{t=0}^T\gamma^t (r^e_t+r^i_t)\right].
	\end{align}
	According to \cite{strehl2008analysis}, \lm{lemma:1} provides a theoretical guarantee on that the policy of an agent will converge to suboptimal with a finite-time bound if the intrinsic bonus is composed by visitation counts as follows, where the bonus encourages the agent to explore the environment in order to reduce the uncertainty.
	
	\begin{lemma}\label{lemma:1}
		The procedure of solving the following augmented Bellman equation for state value is guaranteed to converge to a suboptimality with a finite-time bound
		\vspace{-2pt}
		\begin{equation}
		V(s_t)\!=\!\max_{a_t\in\mathcal{A}}\!\left[\hat{r}(s_t,a_t)\!+\!\gamma \mathbb{E}_{\mathcal{\hat{P}}}V(s_{t+1})\!+\!\frac{\beta}{\sqrt{N(s_{t+1})}}\! \right],
		\end{equation}
		where $\hat{r}$ is the empirical extrinsic reward function, $\hat{\mathcal{P}}$ is the empirical transition function, $\beta$ is a constant, $s_{t+1}$ is the next state of $s_t$ after taking action $a_t$ at step $t$, and $N(s)$ is the visitation count of $s$.
	\end{lemma}
	
	However, keeping exact visit counts are impractical in problems with a large discrete or continuous state space. Previous work like \cite{tang2017exploration} uses an auto-encoder and a hash function to reduce the dimension of state space, but its performance is influenced by hash collisions. Some work employs a density model like PixelCNN to compute the pseudo-counts \cite{ostrovski2017count}, but restrains states to be pixels. Other researchers introduce information-theoretical methods into calculating the curiosity with dynamic models of the environment \cite{mohamed2015variational,houthooft2016vime}. More recent works achieving high performance on hard exploration game use prediction error as intrinsic reward \cite{pathak2017curiosity,burda2018exploration}. Different from these approaches which attempt to build a white-box distribution $\rho_\pi(s)$ or a hash function of states with domain knowledge, we intend to directly estimate $P(\mathbf{1}(s\mbox{ is novel})|s)$ in a generative adversarial fashion.

	\begin{figure}[t]
		\centering
		\includegraphics[width=1\columnwidth]{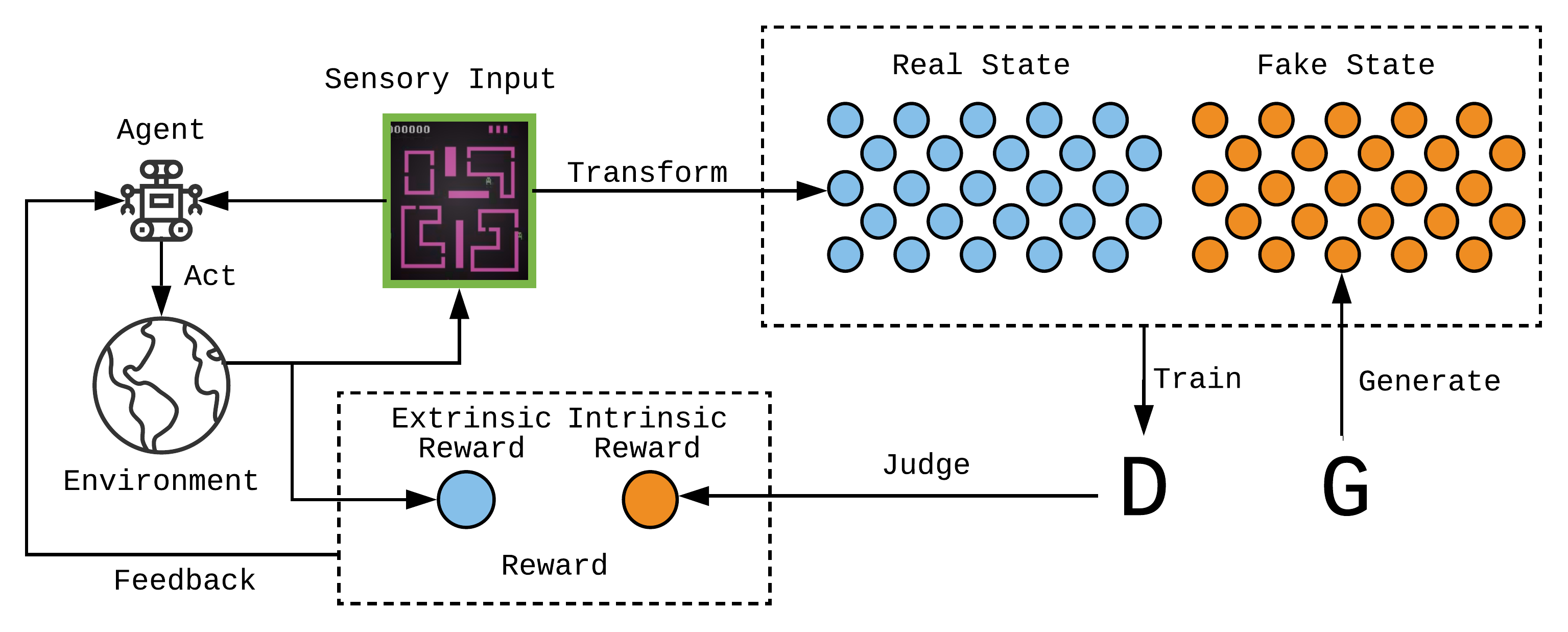}
		\vspace{-15pt}
		\caption{GAEX architecture illustration. $D$ is trained over
			the real state from an abstract feature space and the fake state generated by $G$. The agent learns to act by extrinsic reward received from the environment and intrinsic reward estimated by $D$.}\label{fig:gae_architecture}
		\vspace{-10pt}
	\end{figure}
	
	\subsection{The GAEX Architecture}
	Given a policy $\pi$, according to \cite{kakade2002approximately}, the normalized discounted visitation frequency is written as
	\begin{equation}
	\rho_\pi(s) = (1 - \gamma) \sum \limits_{t=0}^{\infty} \gamma^t P(s_t = s | \pi, s_0),
	\end{equation}
	where actions are selected following the policy $\pi$ and the state transits accordingly. To estimate the curiosity of state $s$, we build negative samples that follow the distribution $g(s)$, and then the probability to be a novel state for state $s$ is given by
	\begin{equation}
	\begin{aligned}
	P(\mathbf{1}(s\mbox{ is novel})|s) &= 1 - P(\mathbf{1}(s\mbox{ is visited})|s) \\
	&= 1 - \frac{\rho_\pi(s)}{\rho_\pi(s) + g(s)}.
	\end{aligned}
	\end{equation}
	
	Therefore, we train a discriminator to represent the probability $D_{\theta_D}(s)=\rho_\pi(s)/(\rho_\pi(s) + g(s))$ indicating the state novelty given state $s$. Additionally, we also train a generator $G_{\theta_G}$ in order to learn the probability $G_{\theta_G}(s)=g(s)$ representing the distribution of negative samples. An illustration of GAEX architecture is shown in Figure~\ref{fig:gae_architecture}.
	
	In GAEX, the generator $G$ is fed with random noise, aiming to generate states as real as they are sampled from the policy interacting with the environment $\rho_\pi(s)$. The discriminator $D$ aims to discriminate between the real states sampled from $\rho_\pi(s)$ and the fake states which are generated from $G$. 
	As GAEX learns, once the agent encounters a novel state $s_{t+1}$ after taking some action $a_t$, $D$ will regard it as a fake state with a low $P(\mathbf{1}(s\mbox{ is visited})|s)$, and a large bonus intrinsic reward $r^i_t$ will be assigned to the novel $s_{t+1}$.

	Note that if $D$ learns in an online manner where real states are sampled sequentially from the environment, GAEX will be risked being affected by the high correlation of recent states, and forget the states it has seen before. Thus it is practically effective to employ an additional experience replay buffer $\mathit{M}$. 
	In particular, \alg{alg:dqn_gae} in the appendix shows the DQN-GAEX algorithm which applies the DQN algorithm into our GAEX framework in detail. For simplicity, we employ the original GAN loss \cite{goodfellow2014generative} to train DQN-GAEX. Moreover, it is worth noting that we take the following techniques within GAEX:

	
	\paragraph{State Abstraction} To efficiently distinguish the real and the fake states, we employ state abstraction to reduce the state space, since the sensory inputs involve too many useless details for distribution modeling. Thus, we transform the original input into a compact feature space $\phi$ which keeps the important and ignores the rest. The detailed transformation process is given in \ap{append:A}.

	\paragraph{Choice of Intrinsic Reward Function $f$} Since the probability of a state given by $D_{\theta_D}(s)$ will increase as similar states have been visited many times, the intrinsic reward should be a monotonic decreasing function of that probability. Our experimental results show the function $f$ with the form $
	f(D_{\theta_D}(s)) = \beta(1-D_{\theta_D}(s))^2$ achieves a great empirical performance, which encourages exploration in a quadratic manner. Here $\beta$ is an adjustable hyperparameter. Other function forms are also open to explore.

	\subsection{Discussions of Training Frequencies}
	
	During our evaluation test, we find a fine-tuned training frequency of GAEX crucial to the performance. In this section, we perform an intuitive analysis from two aspects.
	
	\paragraph{Discriminator and Generator}
	According to common training experience of GAN, $G$ is usually trained more frequently than $D$ to reach the convergence. However, although GAN is used as generating fake data to fit distribution of states with experience, in GAEX the generator should be trained less frequently to make the discriminator provide effective guidance for exploration.
	
	\begin{figure}[t]
		\centering
		\subfigure[]{\includegraphics[width=.33\linewidth]{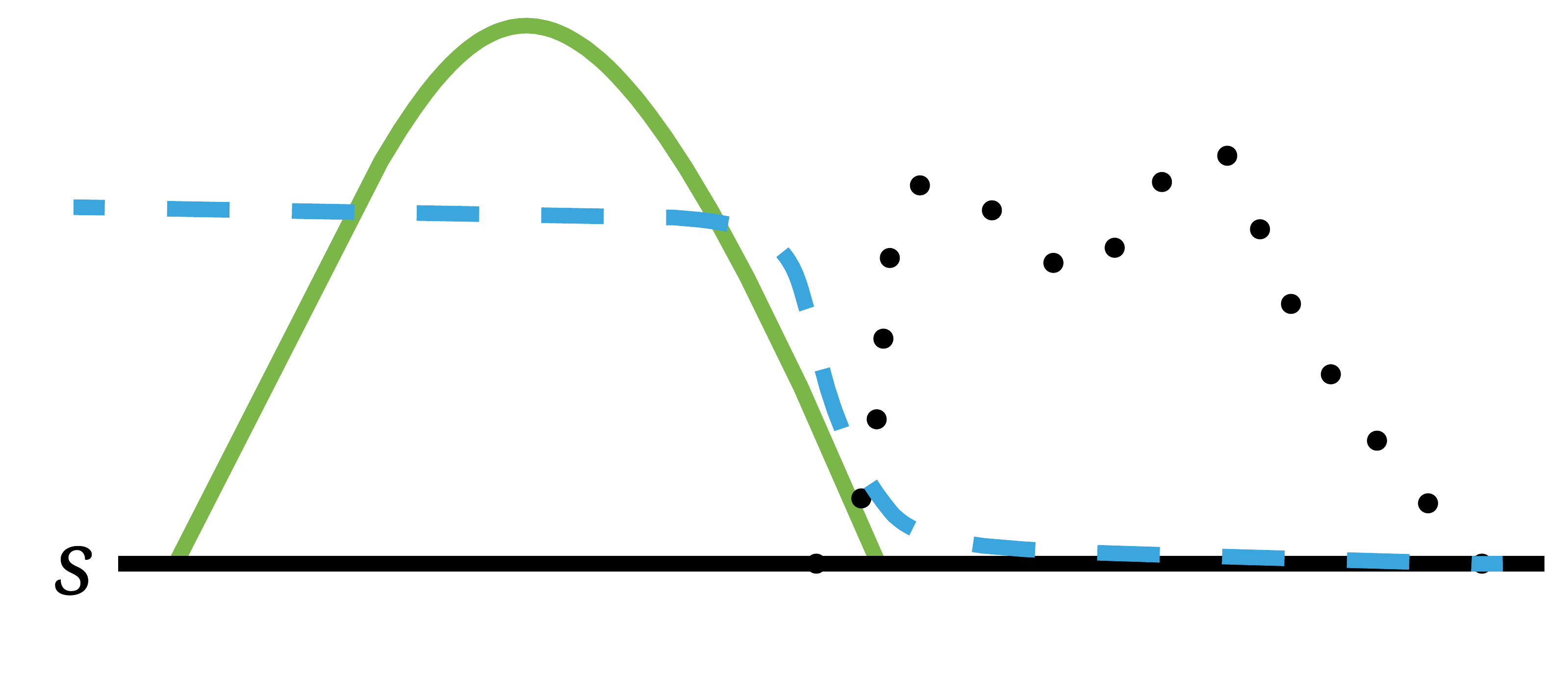}}
		\hspace{-6pt}
		\subfigure[]{\includegraphics[width=.33\linewidth]{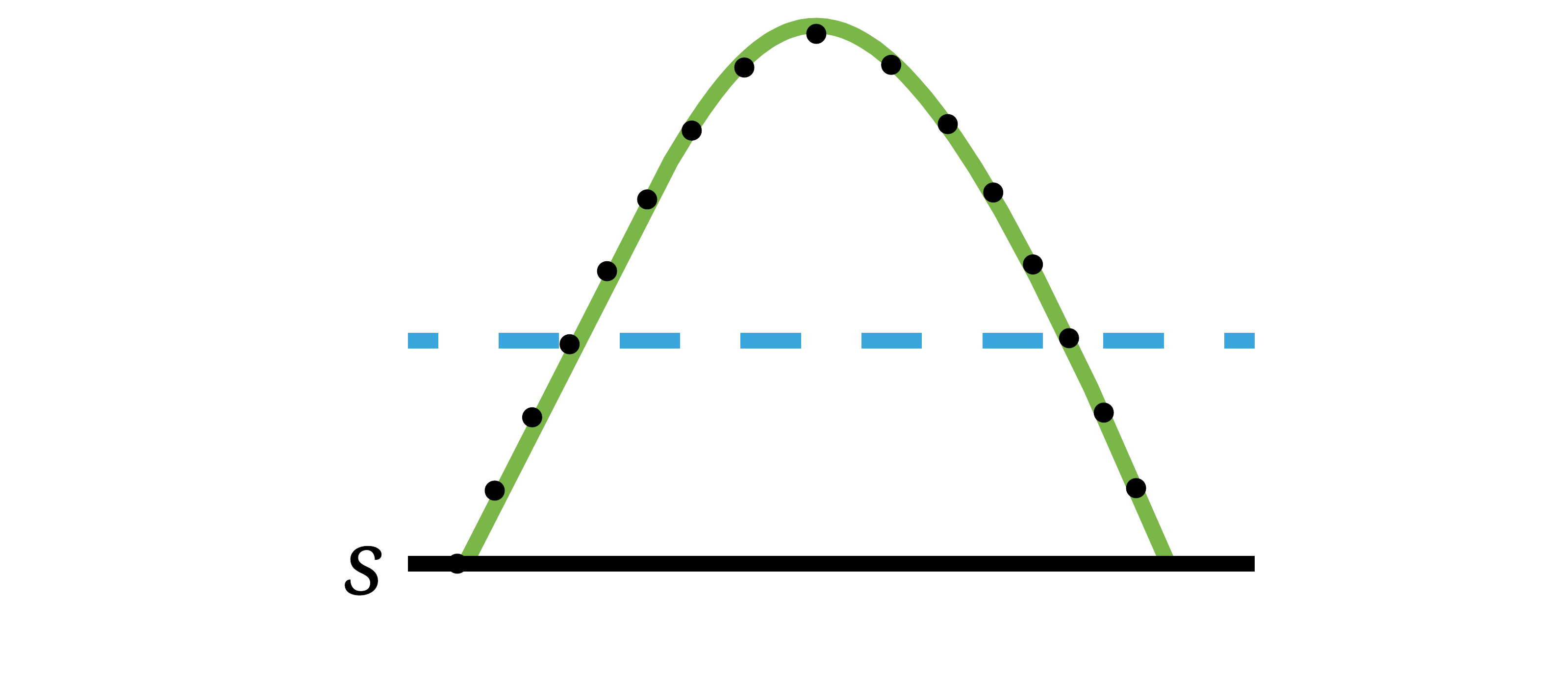}}
		\hspace{-14pt}
		\subfigure[]{\includegraphics[width=.33\linewidth]{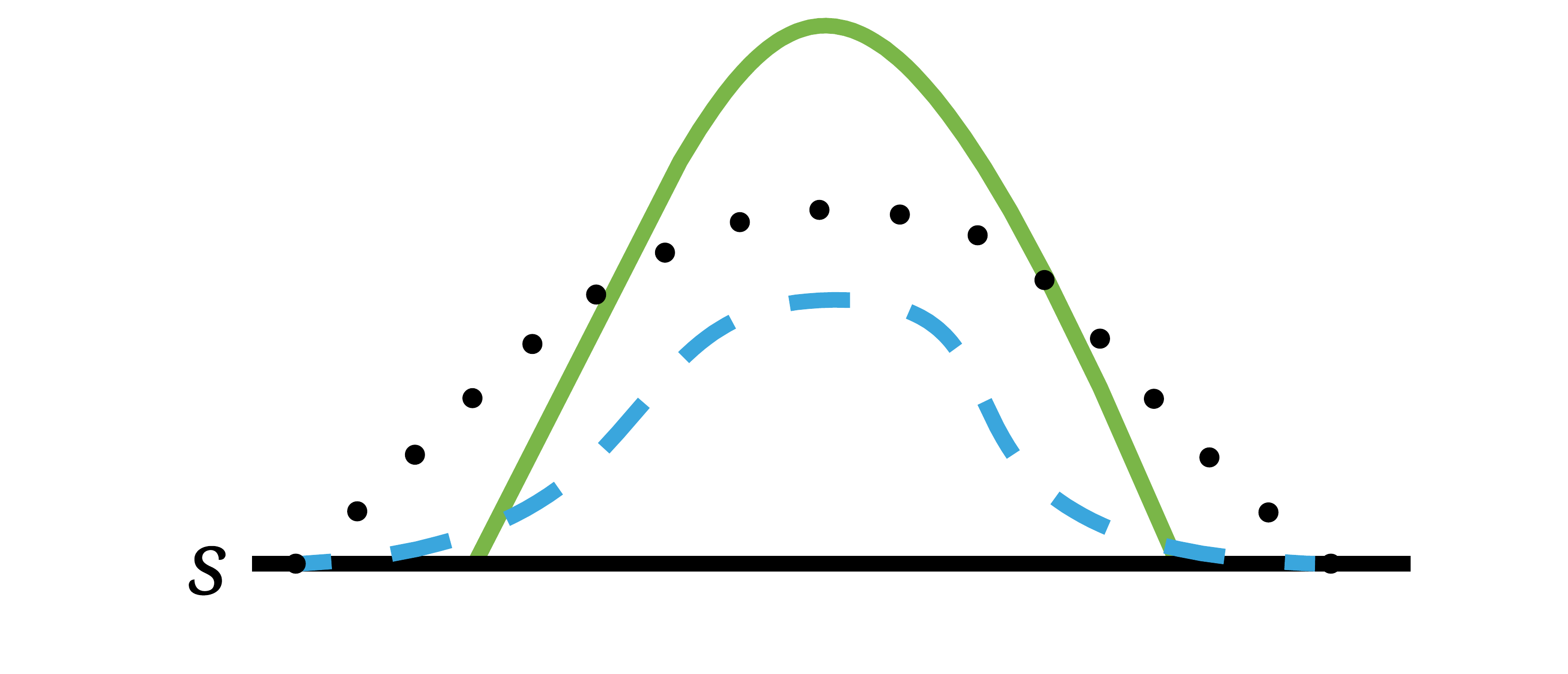}}
		\vspace{-10pt}
		\caption{\small{Three different situations of the training of GAN in which the discriminative distribution $P(\mathbf{1}(s\mbox{ is visited})|s)$ (blue, dashed line) discriminates samples from the fake state distribution $g(s)$ (black, dotted line) and the current states density $\rho_\pi$ (green, solid line). (a) $g(s)$ is far away from $\rho_\pi(s)$, the discriminator is easy to distinguish fake states and occasional visited states. (b) $g(s)$ converges to $\rho_\pi(s)$, states are equally treated, novel states are not encouraged to be explored. (c) $g(s)$ is similar to $\rho_\pi(s)$, the discriminator is fooled by fakes and rarely visited states.
		}}\label{fig:frequency}
	\end{figure}
	
	
	
	To understand how training frequency takes effect, let us think about three situations as shown in \fig{fig:frequency}. Denote $g(s)$ and $\rho_\pi(s)$ the same as in the previous section. When $G$ is not well-trained, leaving $g(s)$ far from $\rho_\pi(s)$, makes $D$ too sharp (output either 1 or 0) to provide effective intrinsic reward to guide the exploration.
	On the contrary, when $g(s)$ is trained well and converges exactly to $\rho_\pi(s)$, states are equally treated, resulting in no encouragement for novel or less visited states. Therefore, it is better to hold $g(s)$ near $\rho_\pi(s)$, where $D$ may be fooled by rarely visited ones, which leads to effective exploration bonus for such states.
	
	Specifically, we set the training frequency of $G$ and $D$ as 1:1 in our experiment. Other frequencies like 1:10 and 10:1 have been tested but they fail to work at all, even in the simplest chain MDP environment, which supports our intuition as shown in \fig{fig:frequency}.

	\paragraph{GAN and Agent}\label{minisection:intuition}
	As the experience data for training GAN is sampled from the current policy of the agent, the state distribution keeps changing along with the policy $\pi$ changes. With this point of view, if we train GAN as frequently as the agent (e.g. the DQN in \alg{alg:dqn_gae}), the policy will change little, so as the state distribution. Thus the generator will learn the distribution fast and discriminator can only provide undifferentiated exploration encouragement as in \fig{fig:frequency}(b). As a result, the GAN's training frequency should be much lower than the agent's training frequency, also leading the computational cost to be largely reduced. 

	\section{Experiment}
	
	In order to test GAEX's ability to guide exploration, we qualitatively and quantitatively evaluate the DQN-GAEX algorithm in different scenarios. Environmental descriptions and training details are also given in this section.
	
	\subsection{Experimental Setup}
	\begin{figure}[t]
		\centering
		\includegraphics[width=1\columnwidth]{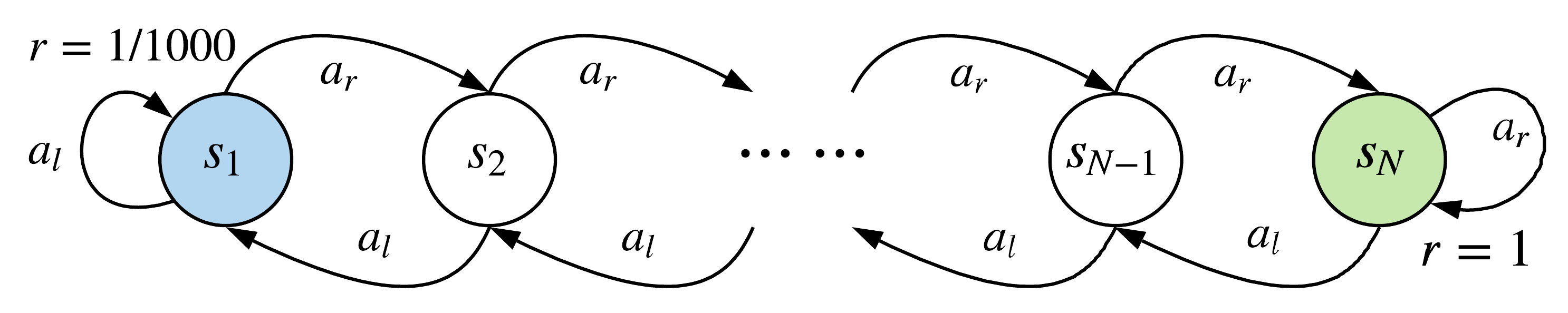}
		\caption{$N$ states Chain MDP environment consists of states $\{s_1, s_2,...,s_N\}$, in which $a_l$, $a_r$ denote the action of moving left and right, respectively. The agent receives reward $r=1/1000$ when it stays in $s_1$ and $r=1$ when stays in $s_N$.}\label{fig:chain_mdp}
	\end{figure}
	The first environment is chain MDP, proposed by ~\cite{osband2016deep} which is designed to be hard to explore. This environment consists of $N$ states $\{s_1, s_2,...,s_N\}$ in a series. The agent always starts in state $s_2$, and can move either right or left in each step, resulting in deterministic transitions to another state except in $s_1$ and $s_N$. The agent can only receive two kinds of rewards: an easily achieved small reward $r=1/1000$ when choose to stay in $s_1$, and a large but hard achieved reward $r=1$ if stay in $s_N$. Each episode contains $N+9$ steps, so an agent can achieve at most a total reward of 10. A uniformly random exploration strategy like $\epsilon$-greedy will cause the agent easily trapped in the local optimal state $s_1$ and nearly impossible to reach to the optimum $s_N$, especially when $N$ is large. 
	
	The second group of environments are Atari games, which have become the deep RL algorithm benchmark. In order to test the performance in sparse reward environments, we follow \cite{bellemare2016unifying} and evaluate our algorithm in the six hard exploration sparse reward environments, i.e. \textit{Private Eye}, \textit{Solaris}, \textit{Venture}, \textit{Montezuma's Revenge}, \textit{Gravitar} and \textit{Freeway}, which are all accessible as parts of {\em OpenAI Gym} \cite{brockman2016openai}.
	
	
	The last environment is the classic Nintendo game \textit{Super Mario Bros} which is available in {\em OpenAI Gym environment for Super Mario Bros}~\cite{gym-super-mario-bros}. In this game, the agent is born with 3 lives at the left of the screen, aiming to complete each stage by moving right to reach the flag pole. Events as touching the monsters and falling into the pits will result in a death, which bring back the game into the beginning or the check point. The main difficulty of this game is that the action space consists of several simultaneous button presses, resulting in complex policy required in some scenarios, e.g. policy that makes a long jump to get over wide gaps or tall tubes. 
	As a final test, we evaluate our algorithm in this environment with no extrinsic reward to see the agent's intrinsic motivation.
	
	Besides the environmental setups, we present the detailed network architectures, hyperparameter settings and training details of the three environments in \ap{append:B}.

	\subsection{Results Analysis}
	We first test DQN-GAEX in a simple chain MDP environment to just show its significant effectiveness. Although the generated states are only used in GAN, our experiment presents the existence of generator is the key to activate exploration. Then experiment on Atari games indicates that our method generalizes well to harder exploration sparse reward environment. And empirical results on \textit{Super Mario Bros} with intrinsic reward only show the agent's strong curiosity to explore novel states. We also verify the intuition discussed above that a much less frequently trained GAN can lead to better performance.

	\begin{figure}[t]
		\vspace{-16pt}
		\centering
		\subfigure{\includegraphics[width=.49\linewidth]{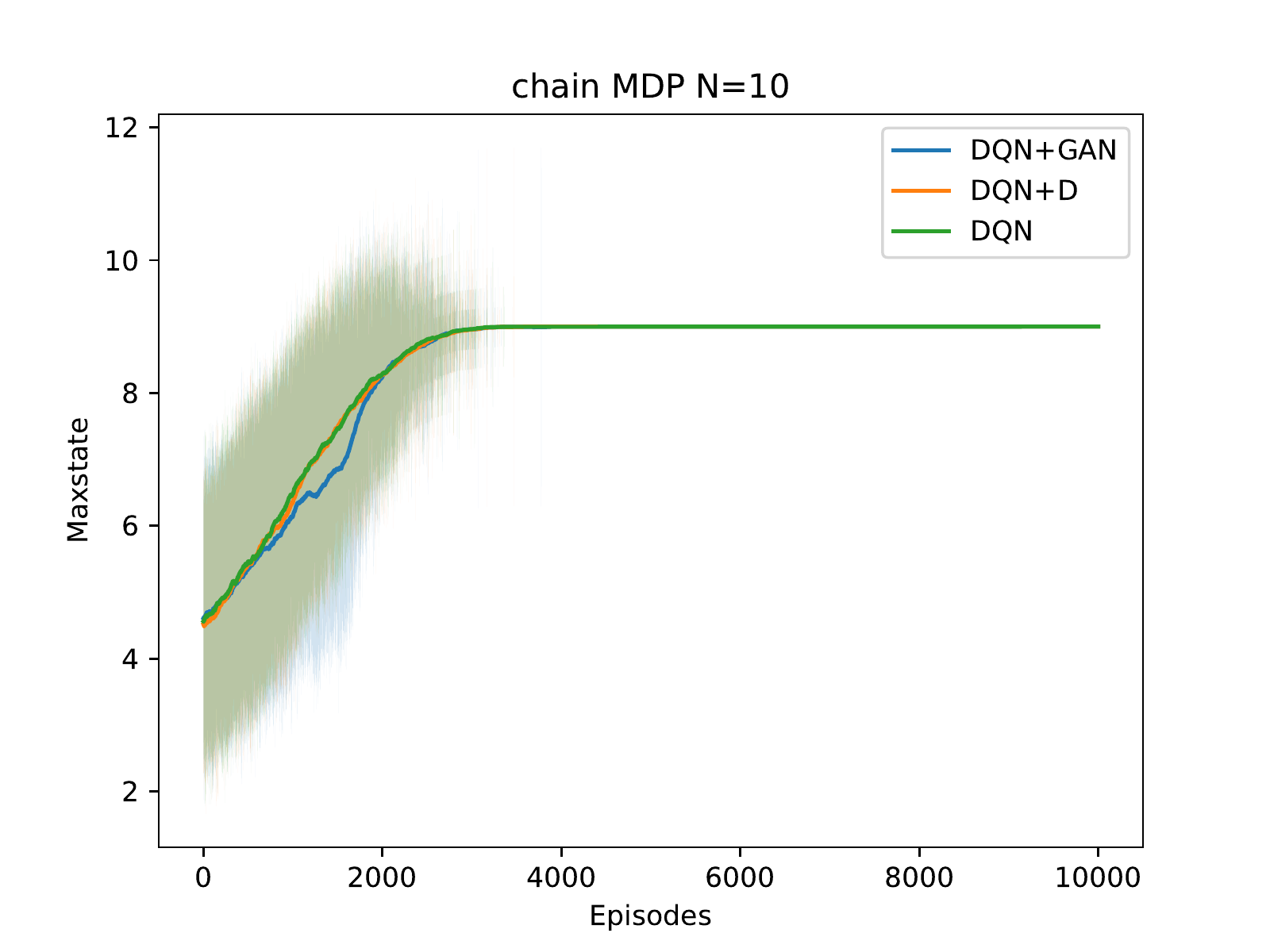}}
		\vspace{-10pt}
		\subfigure{\includegraphics[width=.49\linewidth]{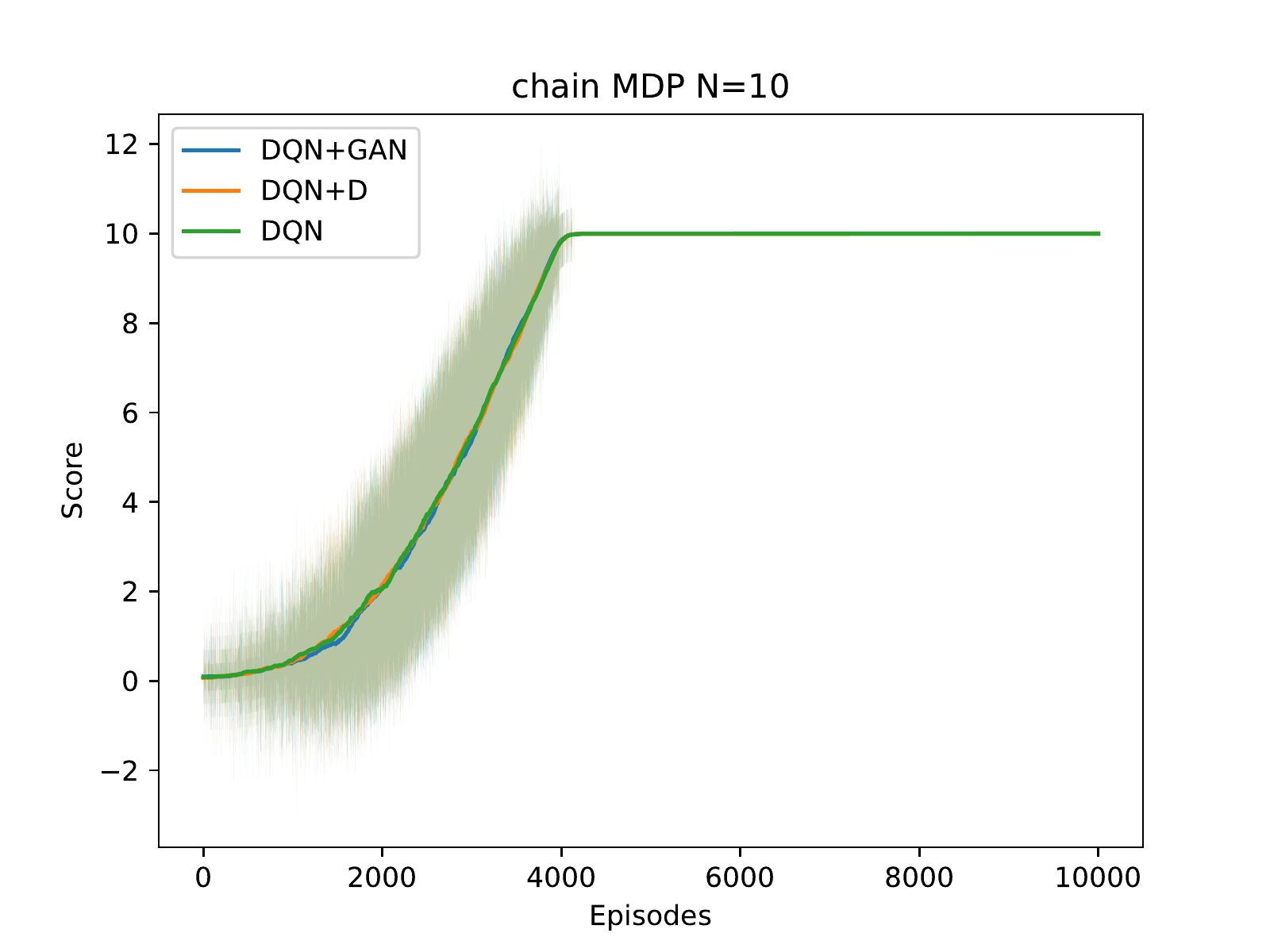}}
		\vspace{-10pt}
		\subfigure{\includegraphics[width=.49\linewidth]{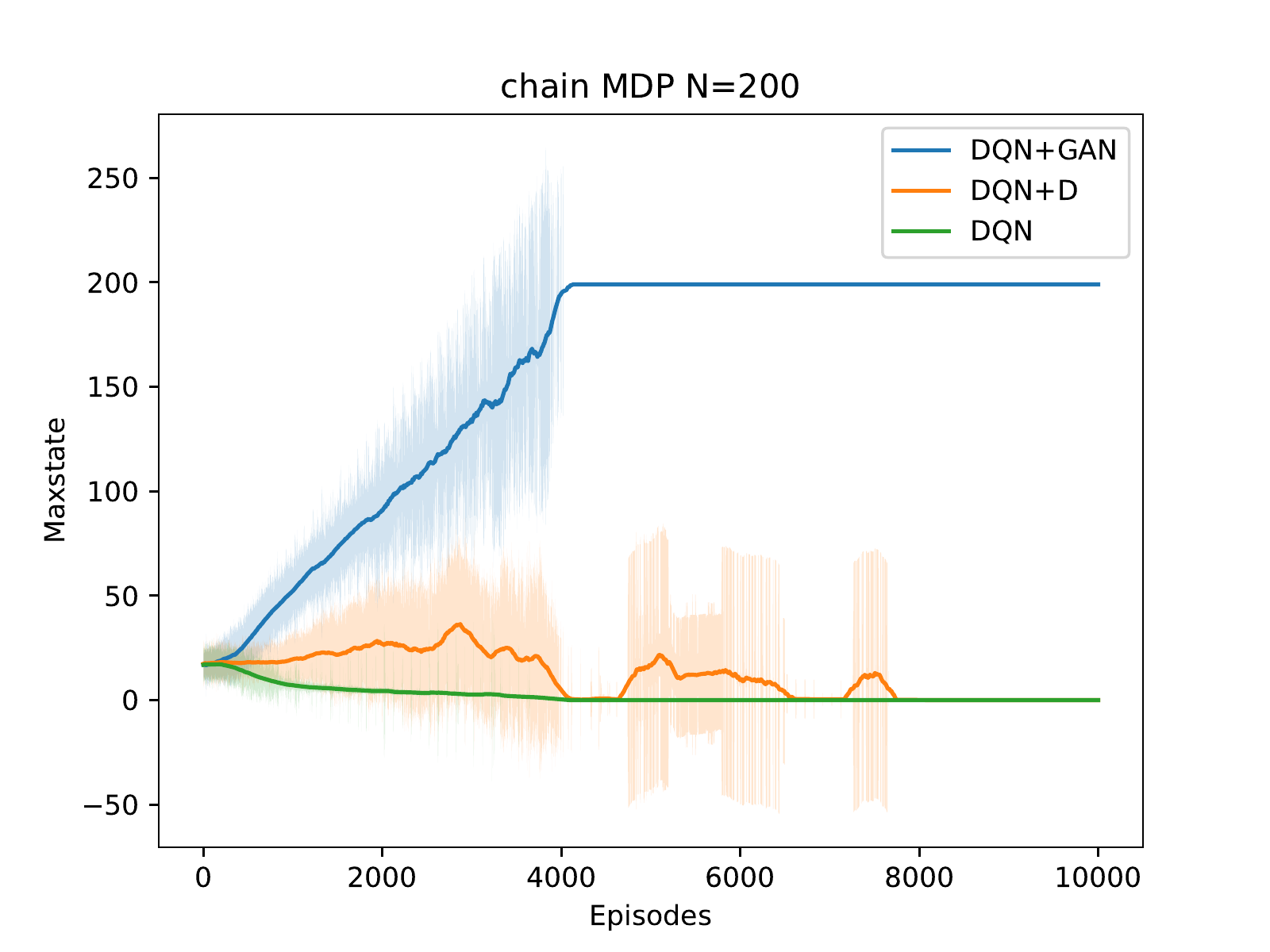}}
		\subfigure{\includegraphics[width=.49\linewidth]{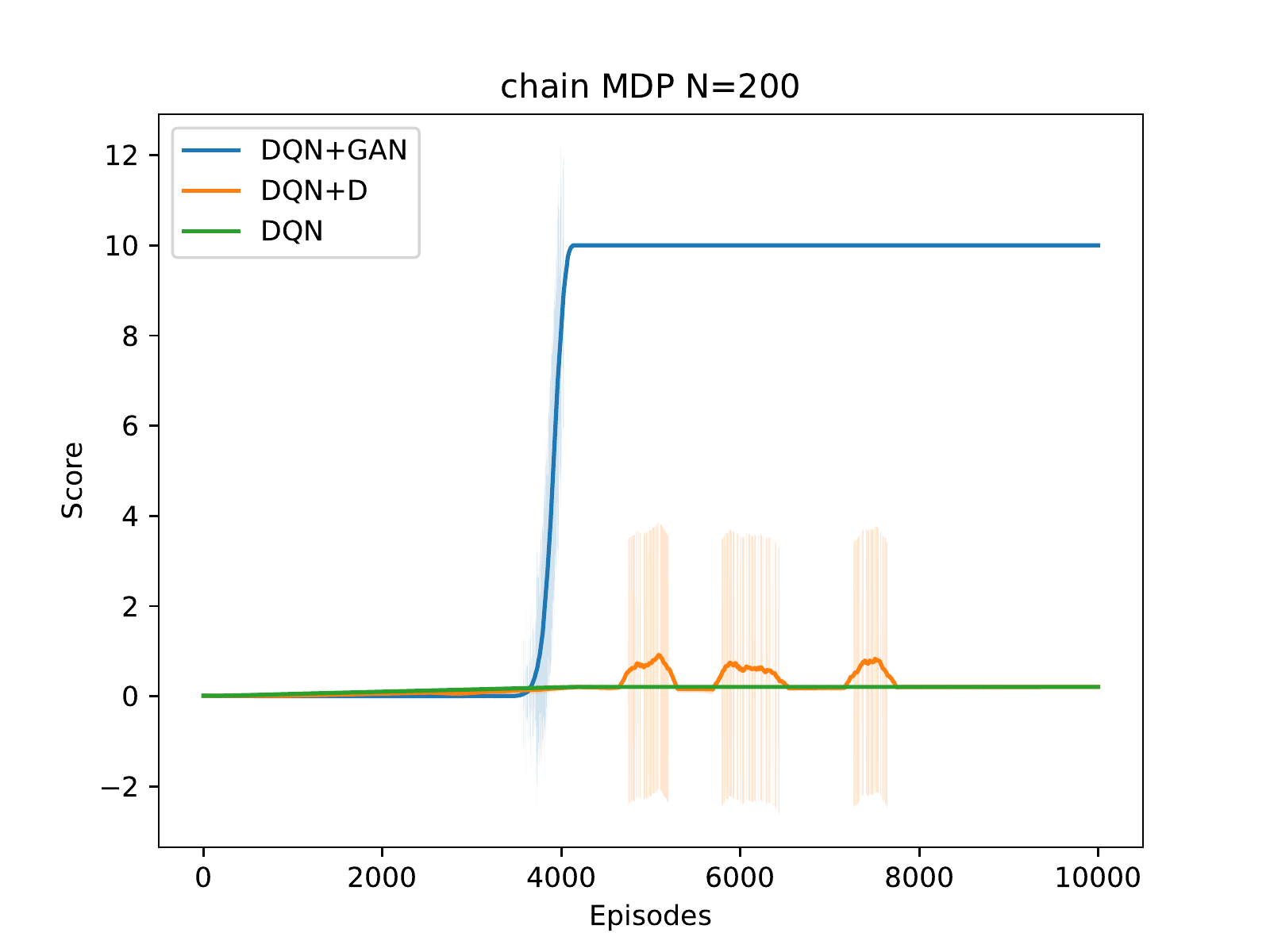}}
		\vspace{-10pt}
		\subfigure{\includegraphics[width=.49\linewidth]{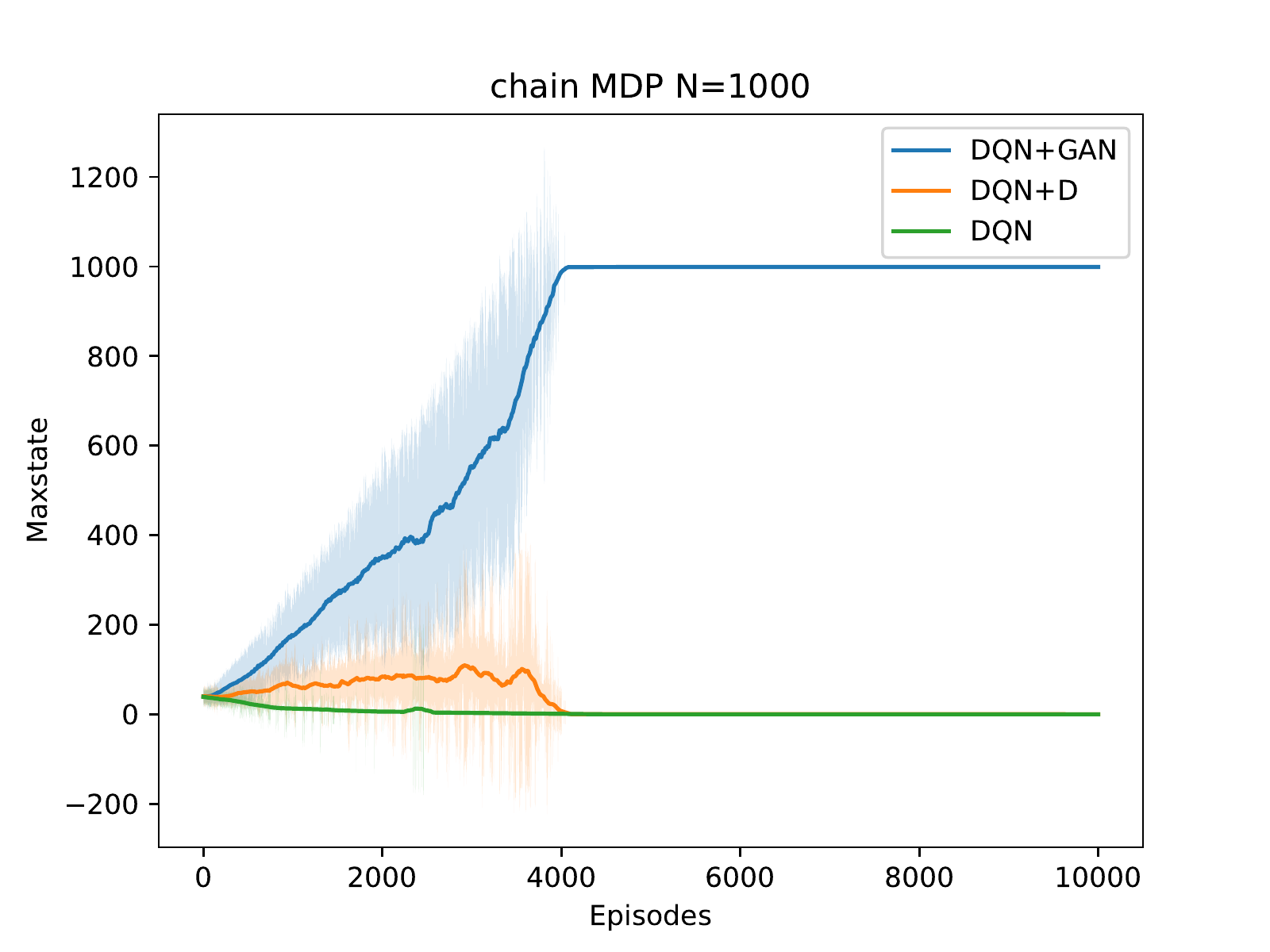}}
		\vspace{-10pt}
		\subfigure{\includegraphics[width=.49\linewidth]{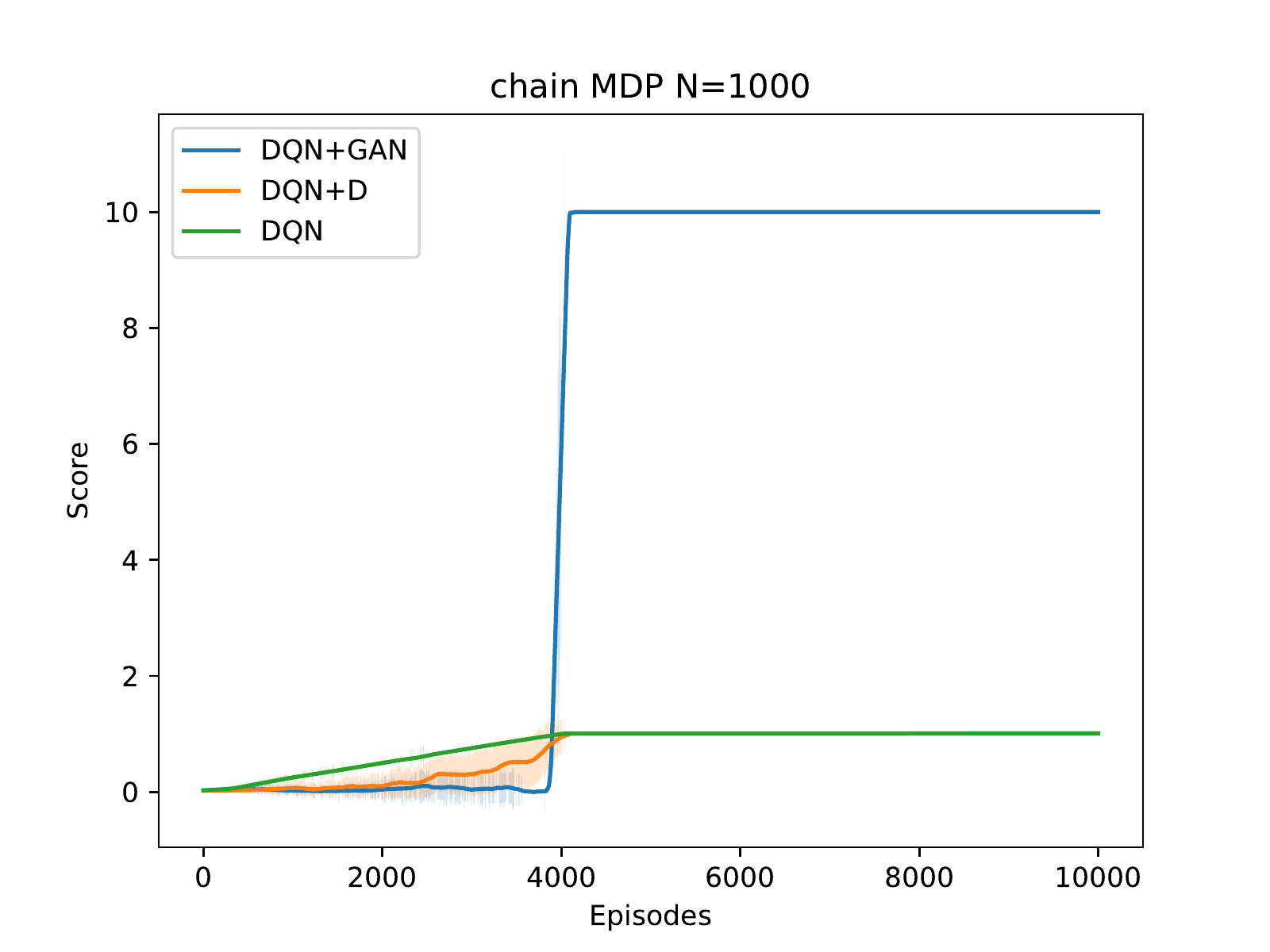}}
		\vspace{12pt}
		\caption{\small{Results of DQN, DQN with a discriminator only (DQN+$D$) and DQN-GAEX (DQN+GAN) in Chain MDP environments with $N = 10$, $N = 200$ and $N = 1000$ states.}}\label{fig:comparasion1}
		\vspace{-16pt}
	\end{figure}

	\paragraph{Verifying Benifits of GAN}
	As shown in \fig{fig:comparasion1}, we compare DQN vs DQN+discriminator (DQN+$D$) vs DQN-GAEX (DQN+GAN) in chain MDP environment with a different chain length $N$. All the results are averaged over 10 runs. The left three figures show the max state that the agent has reached during an episode, and the right three figures show the episodic reward that the agent has achieved. We can see only DQN-GAEX solves all the three tasks.

	When $N=10$, all agents can find out the optimal solution, since in this case even by uniformly random exploration (e.g. when $\epsilon=1$ for $\epsilon$-greedy), the probability of reaching the rightmost state $s_{10}$ from the initial state $s_1$ is about $1/512$, making the environment easy to explore.
	
	Note that the DQN-GAEX agent converges slightly slower than the other two approaches when $N=10$, this is because when $N$ is small, the problem will be quickly solved and most state visitations will be restricted in $s_{10}$, resulting in the less visited states like $s_1$, $s_2$ to be novel. After confirming no more reward can be obtained from these states, the DQN-GAEX agent eventually converges to $s_{10}$. 
	
	When $N=200$ and $N=1000$, DQN gets stuck in $s_1$ from the very beginning, and without the help of generator the DQN+$D$ method can only explore at most 25\% of the chain. Only the DQN-GAEX agent shows active and stable exploration no matter how long the chain is.
	
	\begin{figure}[t]
		\vspace{-16pt}
		\centering
		\subfigure{\includegraphics[width=.49\linewidth]{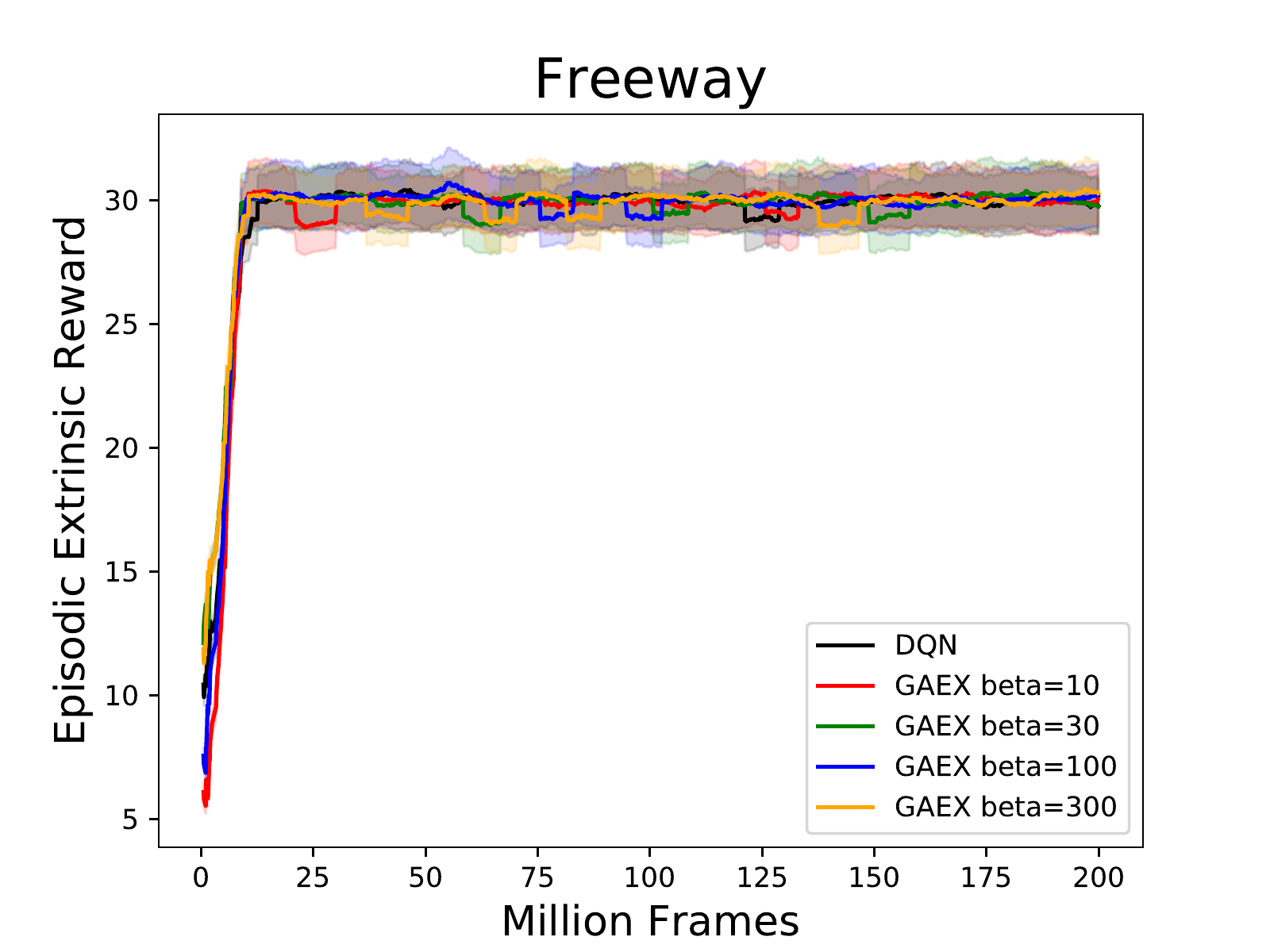}}
		\vspace{-10pt}
		\subfigure{\includegraphics[width=.49\linewidth]{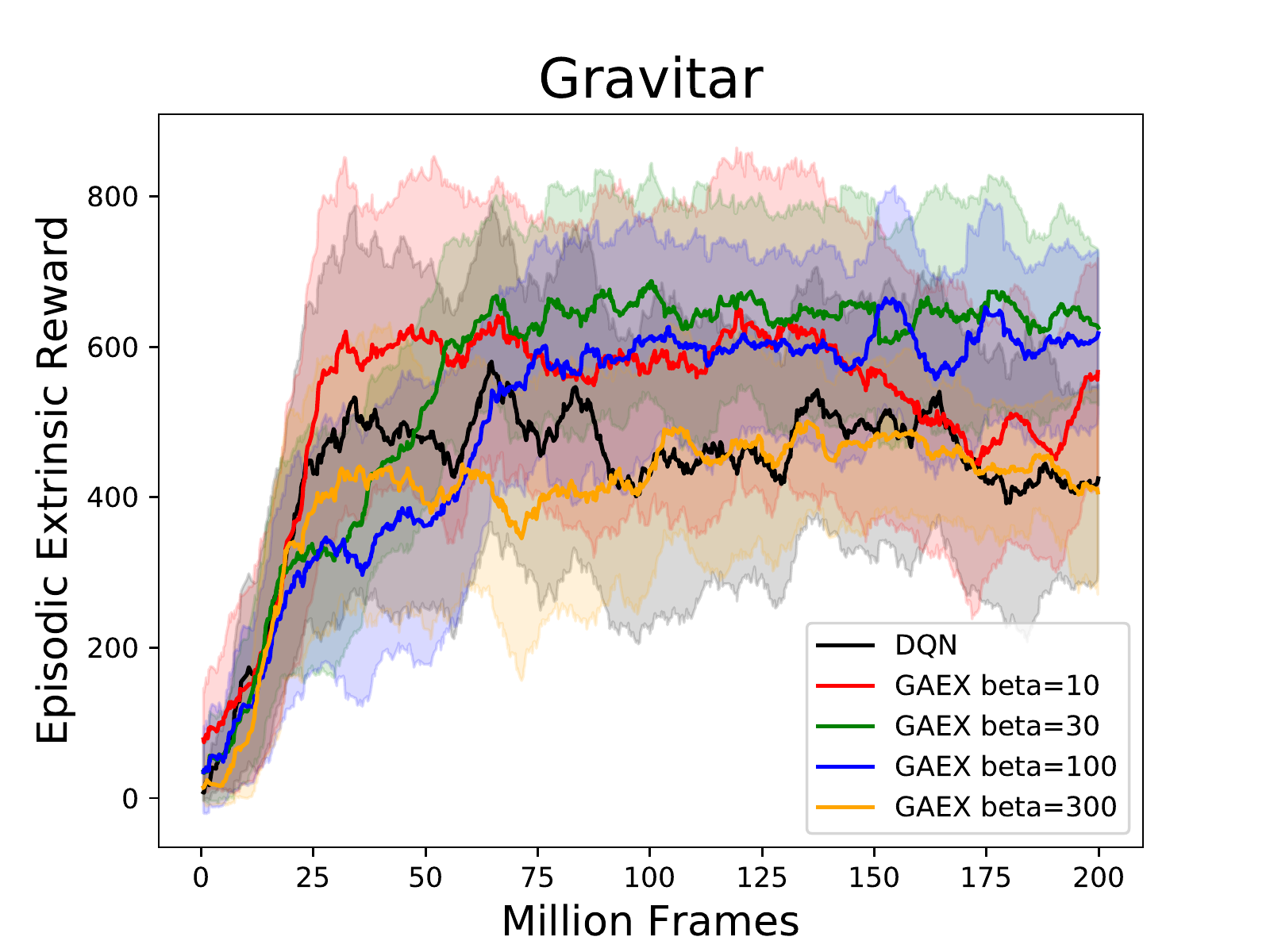}}
		\vspace{-10pt}
		\subfigure{\includegraphics[width=.49\linewidth]{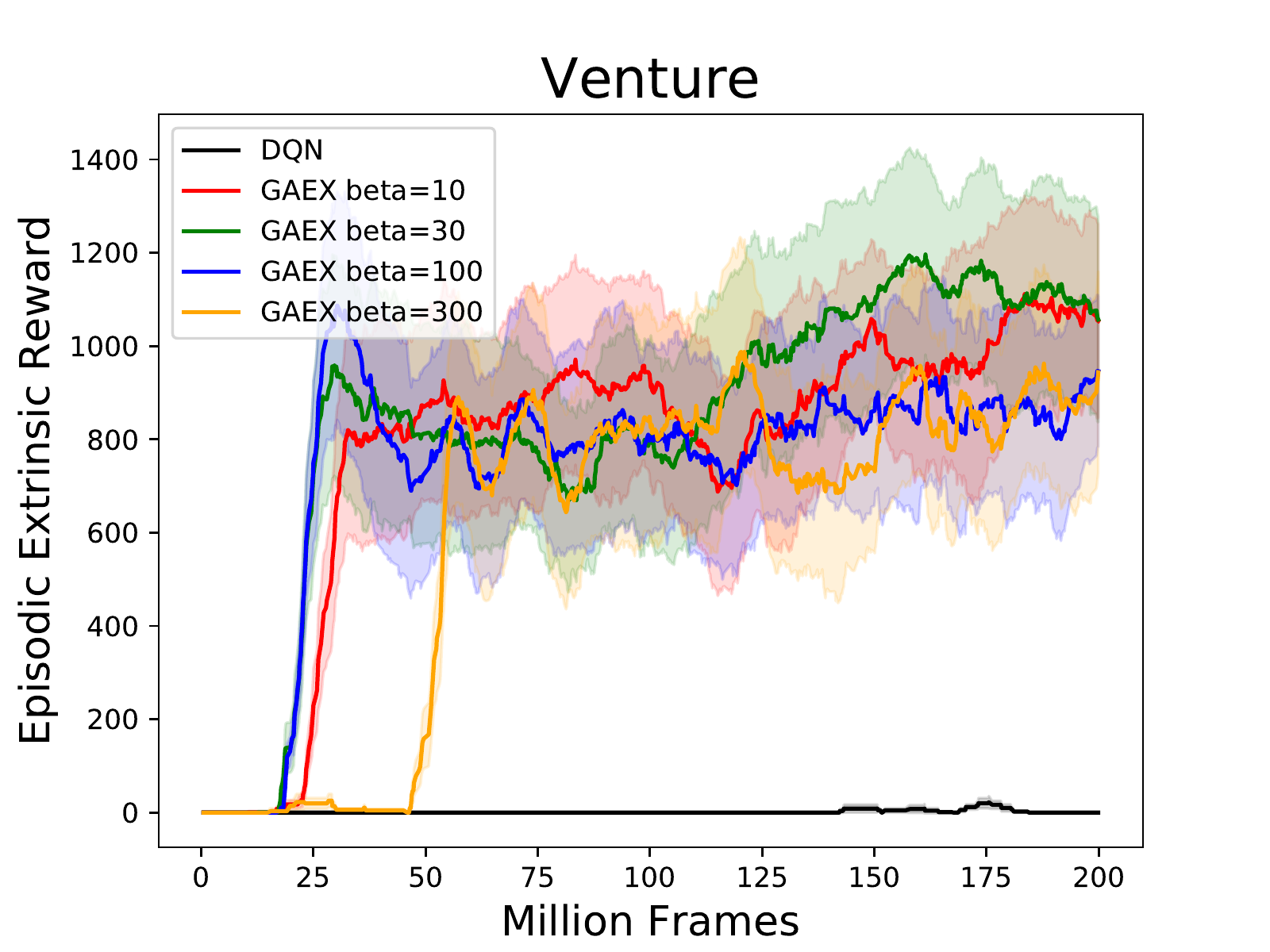}}
		\subfigure{\includegraphics[width=.49\linewidth]{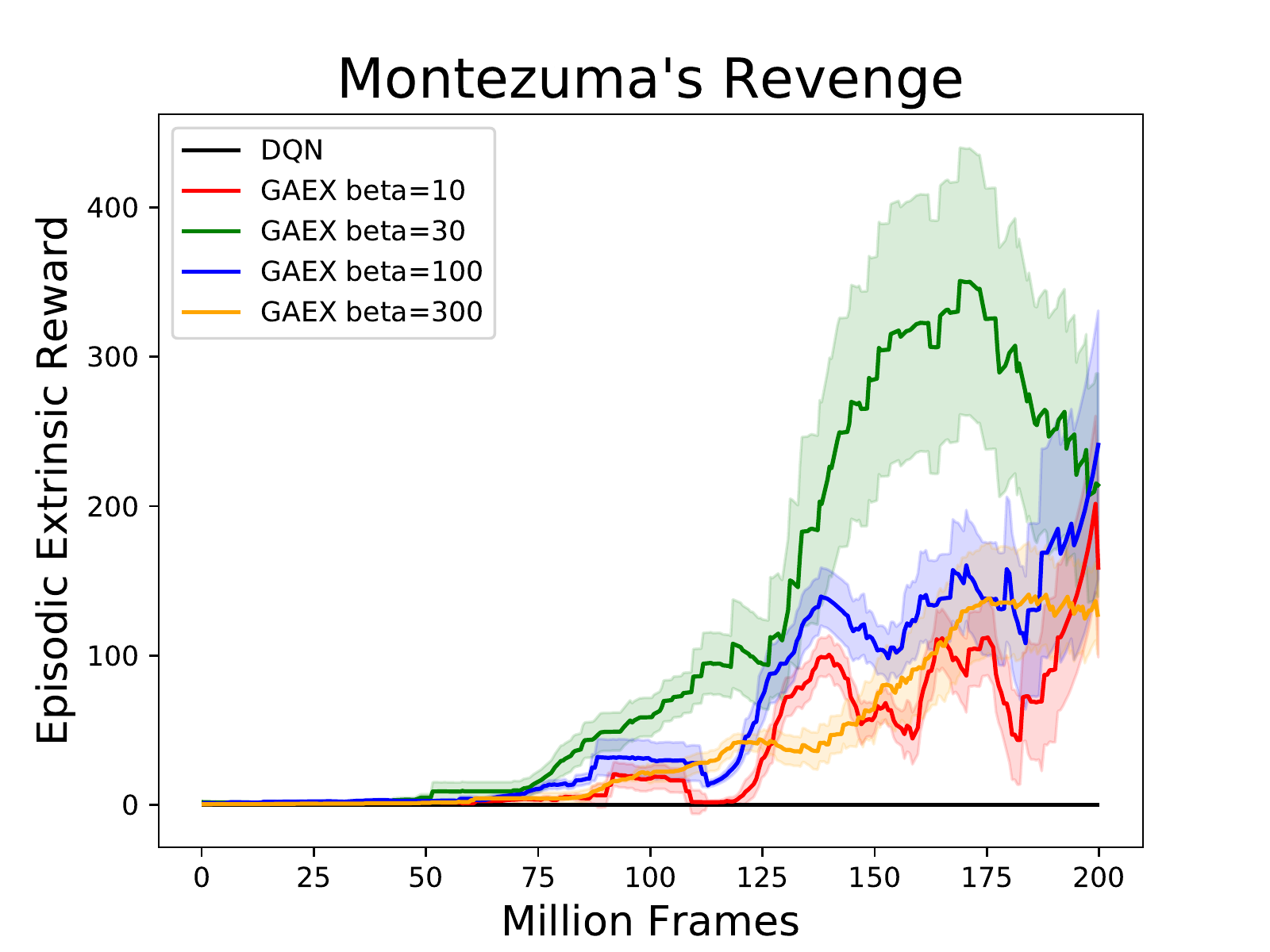}}
		\vspace{-10pt}
		\subfigure{\includegraphics[width=.49\linewidth]{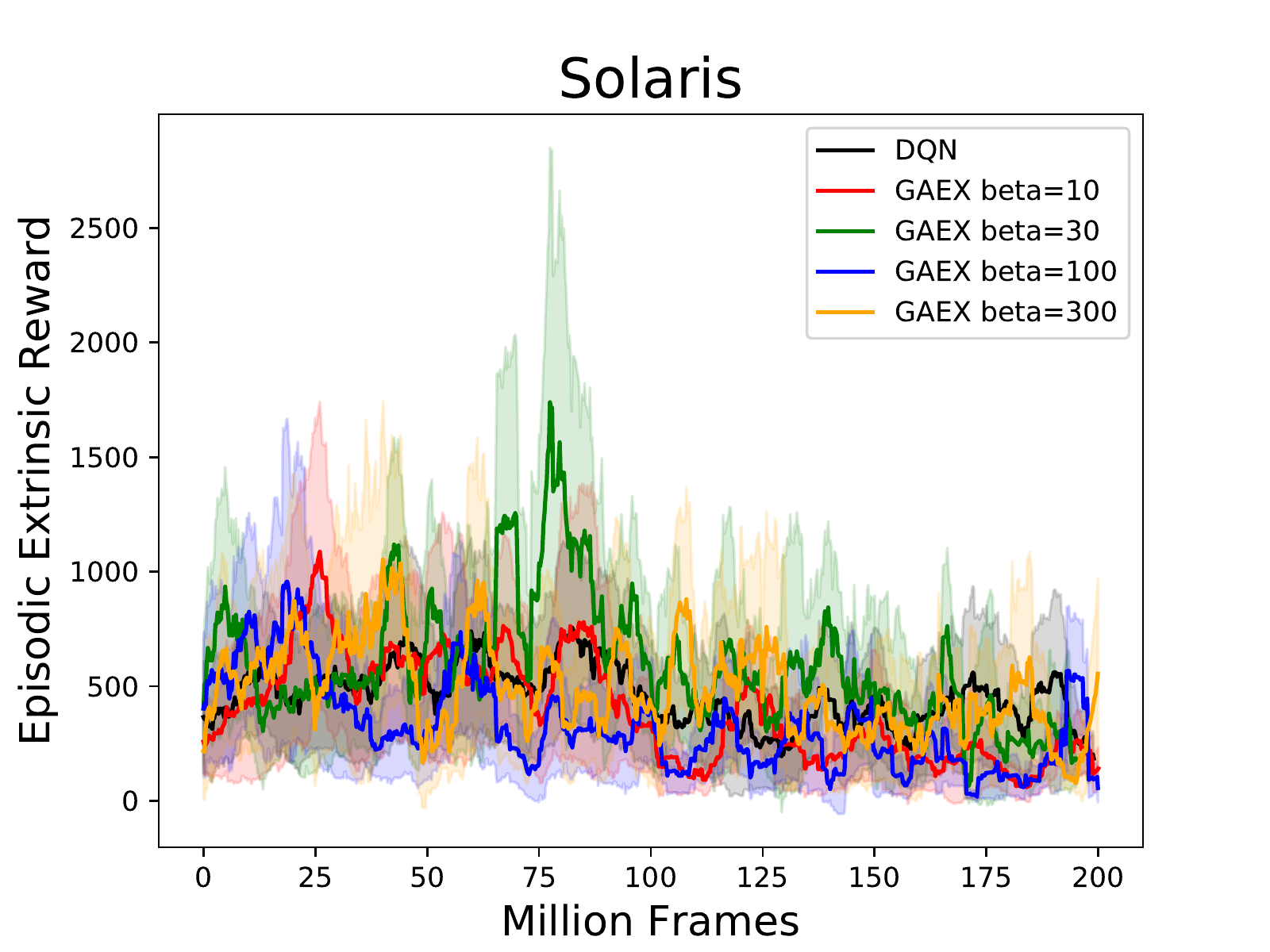}}
		\subfigure{\includegraphics[width=.49\linewidth]{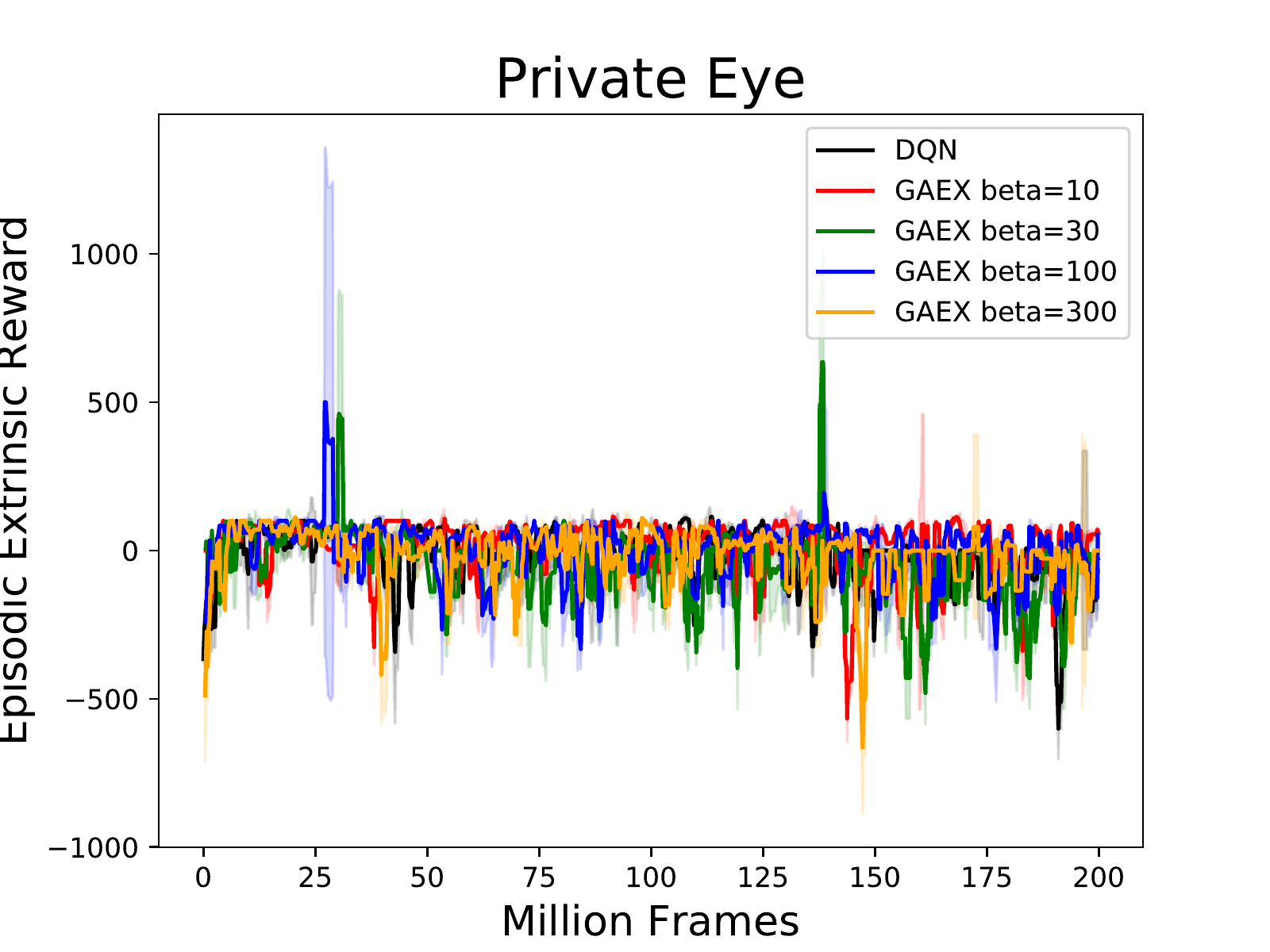}}
		\caption{\small{Compare DQN and DQN-GAEX on various hard exploration sparse reward Atari games. On games \textit{Venture} and \textit{Montezuma's Revenge}, DQN gains no reward at all while GAEX significantly helps the agent explore the environment. }}\label{fig:curves1}
		\vspace{-16pt}
	\end{figure}

	\begin{table*}[!htbp]
		\begin{center}
			\vspace{-7pt}
			{
				\begin{tabular}{ c | c c c c c c | c } 
					Algorithm  & Montezuma & Venture  & Freeway & Solaris & Gravitar & Private Eye & \# Training Frames \\
					\hline
					TRPO-AE-SimHash$^1$ & 75 & 445 & 33.5 & 4467 & 482 & - & 200M \\
					DQN-PixelCNN$^2$ & 2514 & 1356 & 31.7 & \textbf{5502} & 859 & \textbf{15807} & 150M \\
					A2C+CoEX$^3$ & \textbf{6635} & 204 & \textbf{34.0} & - & - & 5316 & 400M \\
					RND (100M frames)$^4$ & 525 & 954 & - & 1270 & 790 & 61 & 100M \\
					\hline
					\textbf{DQN-GAEX (Ours)} & 80 & 1520 & 32.6 & 4692 & 1120 & 2600 & 100M \\
					\textbf{DQN-GAEX (Ours)} & 420 & \textbf{1540} & 32.8 & 4692 & \textbf{1260} & 3782 & 200M \\
					\hline
					Average Human & 4753 & 1188 & 29.6 & 12327 & 3351 & 69571 & / \\
				\end{tabular}
			}
			\caption[]{Comparative evaluations of trained agents ($\beta=30$) on six hard exploration sparse reward Atari games, without further fine-tuning on more effective learning algorithms, only with DQN can we get such a good performance. Baseline results above are taken from: 1) \citet{tang2017exploration} 2) \citet{ostrovski2017count} 3) \citet{choi2018contingency} and 4) \citet{burda2018exploration}.}\label{tab:scores}
			\vspace{-16pt}
		\end{center}
	\end{table*}
	\paragraph{Sparse Reward Scenarios}
	We next evaluate GAEX in six hard exploration sparse reward environments, with different choices of $\beta$. We report the results in a similar way as \cite{ostrovski2017count}, i.e. report the maximum scores averaged over 5 seeds over 200M frames of training, shown in \tb{tab:scores}.
	
	Overall, we observe that the choice of $\beta=30$ performs well on all the six games. We can see that the DQN-GAEX agent significantly outperforms the baseline methods in the game \textit{Venture}, where the score is beyond human average. In \textit{Montezuma's Revenge}, to improve sampling efficiency we replay the trajectory once the agent achieves a higher score, and at last DQN-GAEX can also largely outperforms the baseline. In \textit{Gravitar}, the DQN-GAEX consistently outperforms the baseline, but gets some trouble to further explore the environment. In \textit{Freeway} all the previous methods have nearly saturated performance, including the baseline, which is similar to the DQN-GAEX agent. In \textit{Private Eye} and \textit{Solaris}, the DQN-GAEX does not consistently exceed the performance of DQN, but appears to sometimes discover large reward scenarios, which is not shown by DQN.
	
	It is interesting that there is a huge difference between the scores for A2C+CoEX and DQN-GAEX for the game \textit{Montezuma's Revenge} and \textit{Venture}, and the difference reverses between the two games.  This is probably because A2C+CoEX depends on catching the position of the agent, which is easy for the game \textit{Montezuma's Revenge}, but the pixel-size agent in \textit{Venture} is too small to get attention. 
	
	Note that the currently state-of-the-art approach RND \cite{burda2018exploration} achieves a higher score on these games after training with 1.97 billion frames of experience. However, according to Table 3 in \cite{machado2018count}, our approach is more sampling efficient which achieves better results after training with only 100-200 million frames. Without further fine-tuning on other competitive learning methods such as PPO \cite{schulman2017proximal} and A3C \cite{mnih2016asynchronous}, only with DQN can we get such a great performance.


	\begin{figure}[t]
		\vspace{-10pt}
		\centering
		\subfigure{\includegraphics[width=.49\linewidth]{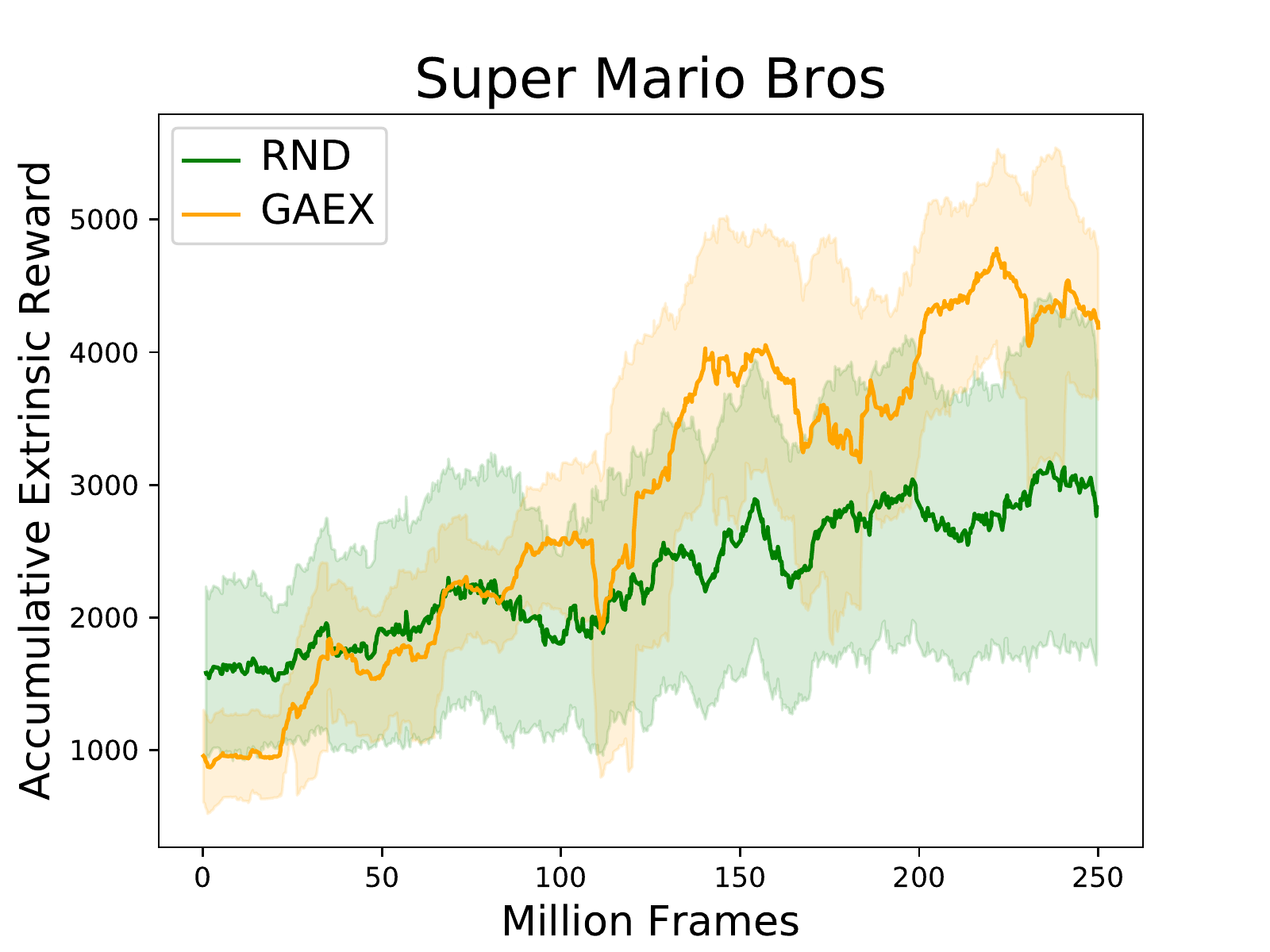}}
		\subfigure{\includegraphics[width=.49\linewidth]{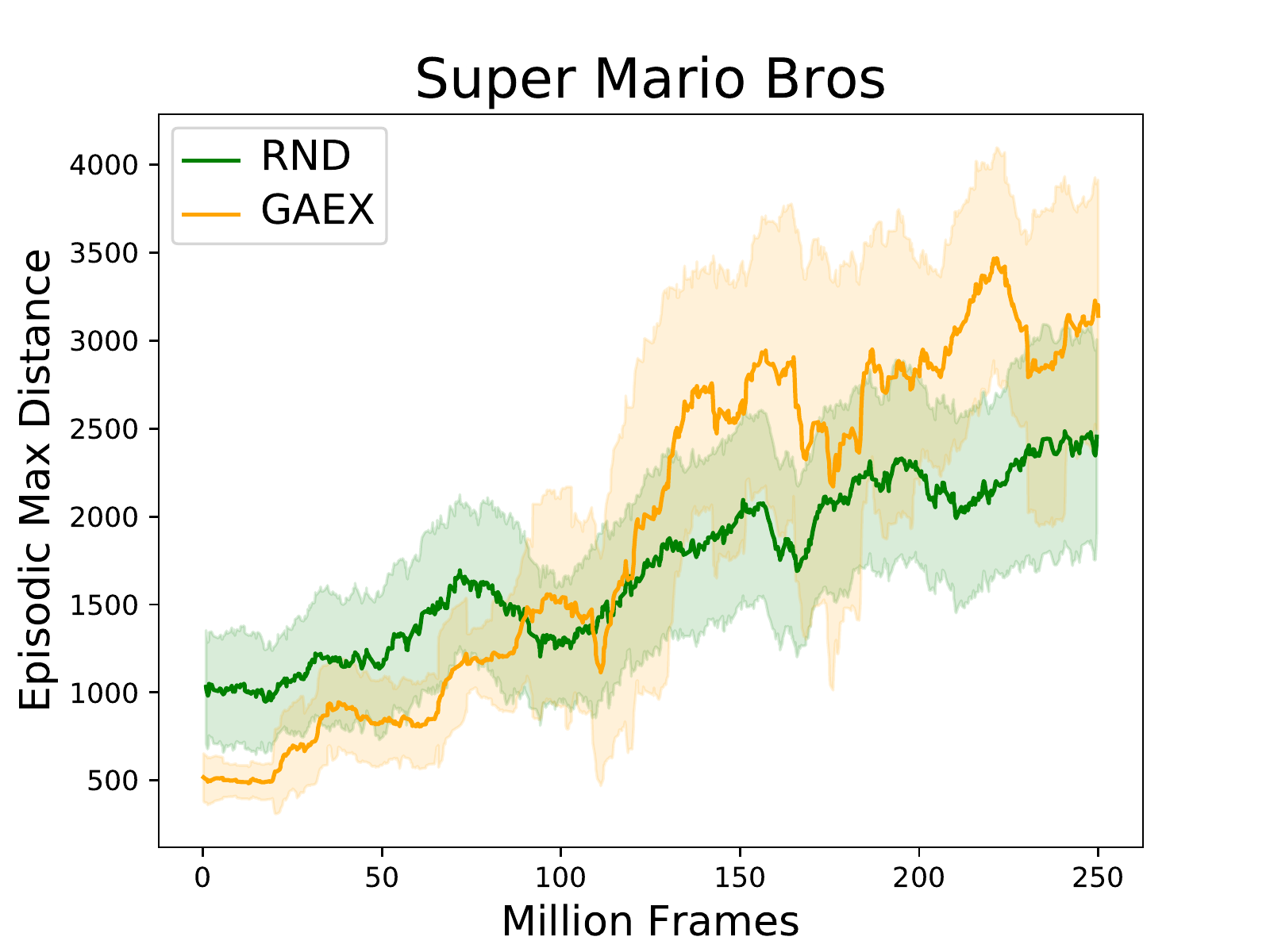}}
		\caption{\small{Compare RND and GAEX on \textit{Super Mario Bros} without external reward signal. The left figure shows the episodic extrinsic reward that the agents can achieve by training with only intrinsic reward, and the right figure show the episodic max distance the agents can reach from the start point. The results are averaged over 3 runs, showing that our GAEX significantly outperforms the RND method in this case.}}\label{fig:mario}
		\vspace{-12pt}
	\end{figure}

	\paragraph{No Reward Setting}
	Environments mentioned above are all of sparse reward setting. However, we are eager to know how GAEX can generalize without extrinsic reward from the environment, where the behavior of the agent is only guided by the intrinsic signal from GAN.
	Therefore, we train DQN-GAEX agent in the complex \textit{Super Mario Bros} environment with no reward to evaluate its generalization ability, since this game shares significant similarities with the previous chain MDP environment, where the agent should keep moving right without any external signal.
	
	To our surprise, the agent can automatically discover meaningful behaviors like jumping over taps or killing enemies to keep alive, which leads to larger intrinsic bonuses, although there is no environmental reward can be received.
	
	In our experiment, after consuming 250M training frames with only the curiosity signal offered by GAN, the agent can successfully completes the level 1-1 and then finishes 40\% of the level 1-2. This result strongly surpasses the previous work A3C+ICM \cite{pathak2017curiosity}, where the agent can just learn to cross 30\% of level 1-1 without extrinsic reward. We also compare our algorithm with the state-of-the-art method RND, and the results are presented in \fig{fig:mario}. Note that we keep a very small exploration rate $\epsilon=0.02$ during the whole training, resulting in a lower performance compared to RND in the very beginning, but GAEX outperforms RND at last, in terms of both the episodic extrinsic reward and the episodic maximum distance from the start point.

\begin{figure}
		\vspace{-10pt}
		\centering
		\includegraphics[width=0.6\columnwidth]{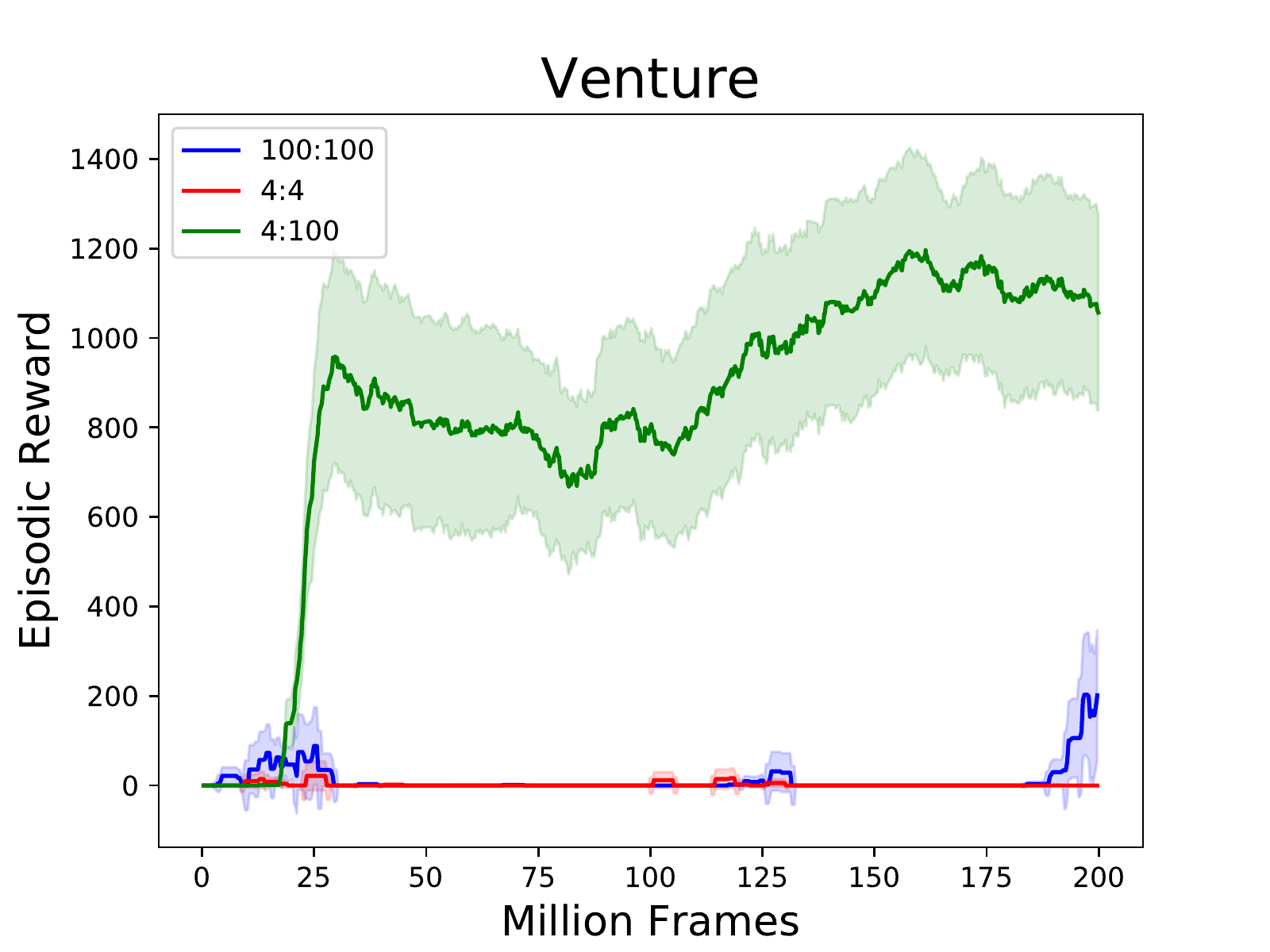}
		\caption{Comparative results on \textit{Venture} under different training frequencies between DQN and GAN, which are denoted in the legend, e.g. the green denotes that DQN is trained for every 4 steps and GAN for every 100 steps, which is the only worked option. Note that we do not evaluate the case when GAN is trained more frequently than the agent because of its extreme inefficiency.}\label{fig:curves2}
		\vspace{-15pt}
	\end{figure}

	\paragraph{Training Frequencies Comparison}\label{minisec:frequence}
	As discussed in \se{minisection:intuition}, the updating frequency between GAN and the agent is critical to the performance. Concretely, GAN should be trained much less frequently than the agent. \fig{fig:curves2} shows the results under different settings of updating frequency of these two parts in \textit{Venture}. As presented, when DQN and GAN are updated with the same frequency (shown as blue and red curves), the agent can hardly learn anything, which supports our insight that the GAN's training frequency should be lower than the agent's.

	\section{Conclusion}
	
	This paper provides a novel exploration framework, Generative Adversarial Exploration (GAEX), to address the exploration-exploitation dilemma in deep reinforcement learning. We apply DQN into our GAEX architecture which leads to the DQN-GAEX algorithm, and employ DQN-GAEX into various exploration scenarios. In the simple chain MDP environment, we confirm the active effect of GAEX on encouraging exploration. In sparse reward and even no reward settings, our method is able to achieve convincing results without further fine-tuning. We finally compare and analyze the training frequency between the agent and GAN. For future work, the GAEX framework can be combined with other general RL algorithms like PPO and SAC to evaluate its performance, as well as other kinds of state abstraction approaches.
	
	\section{Acknowledgments}
	
	We thank the support of CCF-Tencent Open Fund and NSFC (61702327, 61772333, 61632017).
	
	\bibliographystyle{ACM-Reference-Format}
	\bibliography{sample-base}
	
	\clearpage
	\normalsize
	\appendix

	\section{Details of State Abstraction}\label{append:A}
	When the states are represented as raw pixels, it is necessary to employ state abstraction to efficiently distinguish the real states and the fake states. In order to minimize the impact of state abstraction to the stability of training as mush as possible, we design a fixed state abstraction method rather than a learnable approach. Our static state abstraction method is designed in a general way to capture the most information related to the training.
	
	For each of the original state $s_{(j)}'$ (which has already been downsampled and stacked as a $4\times 84 \times 84$ input for DQN), before it is fed into the discriminator, we transform it into a compact feature space by a second-time downsampling procedure as follows:
	
	1) The first $84\times 84$ frame is resized to $8\times 8$ and then flatten as a 64-dimensional vector $\phi_1$.
	
	2) Calculate the row averages and the column averages of the dynamics of the four frames, denoted as $\phi_2$ and $\phi_3$.
	
	3) Construct the state abstraction by concatenating $\phi_1$, $\phi_2$ and $\phi_3$, which is a 148-dimensional vector.
	
	The downsampling procedure largely reduces the information needed to just identify the game stage, and the dynamics extracting procedure makes the GAN focus more on the changing parts of the environment. The reason of taking the averages over columns and rows is that it keeps enough information to figure out the rough position of moving objectives but reduces the dimension. The details of this procedure is listed in \alg{alg:abstraction}.

	\section{Training Details}\label{append:B}
	The detailed experimental setup of all the three environments are shown in \tb{tab:dqn}, \tb{tab:gan} and \tb{tab:atari}. In all the experiments we use the double DQN \cite{van2016deep} with dueling architecture \cite{wang2015dueling}, except for the game \textit{Freeway} since it achieves very bad performance as reported in \cite{wang2015dueling}.
	
	In chain MDP environment, each state is represented as an one-hot vector $\mathbf{1}\{x=s_N\}$. Note that the results shown in \fig{fig:comparasion1} is trained with an equal frequency between DQN and GAN, but this hyperparameter is not as sensitive to the result as in Atari games, we also observe that less frequently trained GAN will lead to similar performance in chain MDP.

	\begin{algorithm}[tb]
		\caption{State Abstraction $\phi$}
		\label{alg:abstraction}
		\flushleft
		\textbf{Input}: State $s_{(j)}'$, which is a list consisting of 4 stacked frames, each of them has the size $84 \times 84$
		\\
		\textbf{Output}: $\phi\left(s_{(j)}'\right)$
		\begin{algorithmic}[1] 
			\STATE rescale the first frame $s_{(j)}'[0]$ into size $8\times 8$, and flatten it as a 64-dimensional vector $\phi_1\left(s_{(j)}'\right)$
			\STATE initialize a placeholder $d\left(s_{(j)}'\right)$ for dynamic features to be a $84\times 84$ zero matrix
			\STATE $\alpha\leftarrow 0.8$
			\FOR {$i=1$ to $3$}
			\STATE $d\left(s_{(j)}'\right) := \alpha d\left(s_{(j)}'\right) + \left(s_{(j)}'[i] - s_{(j)}'[i-1]\right)$
			\ENDFOR
			\STATE Rescale $d\left(s_{(j)}'\right)$ to $42\times 42$
			\STATE $\phi_2\left(s_{(j)}'\right) \leftarrow$ row averages of $d\left(s_{(j)}'\right)$
			\STATE $\phi_3\left(s_{(j)}'\right) \leftarrow$ column averages of $d\left(s_{(j)}'\right)$
			\STATE $\phi\left(s_{(j)}'\right)\leftarrow concate\left(\phi_1\left(s_{(j)}'\right), \phi_2\left(s_{(j)}'\right), \phi_3\left(s_{(j)}'\right)\right) \Big/ 10$ , which is a 148-dimensional vector.
		\end{algorithmic}
	\end{algorithm}

	\begin{table*}[!htbp]
		\begin{center}
			\vspace{17pt}
			{
				\begin{tabular}{c | c | c } 
					& Chain MDP & Atari \& Mario \\
					\hline
					Architecture & MLP: (64, 128, 256, 128) & CNN+MLP (the same as \cite{mnih2015human}) \\
					\hline
					Activation function & ReLU & ReLU \\
					\hline
					Use double DQN & True & True \\
					\hline
					Use dueling architecture & True & True (except for game \textit{Freeway}) \\
					\hline
					Dueling architecture & fully connected & fully connected \\
					\hline
					Optimizer & Adam & RMSProp \\
					\hline
					Learning rate & 0.005 & 0.00025 \\
					\hline
					$n$-step return & 10 & $n$ uniformly sampled from $1, 2, ..., 10$ \\
					\hline
					Update frequency & every 2 episodes  & every 4 steps \\
					\hline
					$\gamma$ & 0.99 & 0.99 \\
					\hline
					Batch size & 300 & 32 \\
					\hline
					Replay memory size & 10000 & 1000000 \\
					\hline
					Gradient clipping & False & [-1,1] \\
					\hline
					Replay start size & 10000 & 50000 \\
					\hline
					Target network update frequency & 50 & 10000 \\
					\hline
					Initial $\epsilon$ & 1.0 & 1.0 (Atari), 0.02 (Mario) \\
					\hline
					Final $\epsilon$ & 0.0 & 0.01 (Atari), 0.02 (Mario) \\
					\hline
					$\epsilon$ decay for every training step & 0.0005 & $10^{-6}$ to $\epsilon=0.1$ and $5\times10^{-10}$ to $\epsilon=0.01$ (Atari)  \\
					\hline
				\end{tabular}
			}
			\caption[]{Hyperparameters of DQN.}\label{tab:dqn}
		\end{center}
	\end{table*}

	\begin{table*}[!htbp]
		\begin{center}
			{
				\begin{tabular}{c | c | c } 
					& Chain MDP & Atari \& Mario \\
					\hline
					Generator architecture & 2 FC hidden layers: (50, 50) & 3 FC hidden layers: (296, 148, 148) \\
					\hline
					Discriminator architecture & 2 FC hidden layers: (50, 50) & 3 FC hidden layers: (148, 74, 74) \\
					\hline
					\# noise & the same as chain length & 128 \\
					\hline
					Activation function & LeakyReLU, $\alpha=0.01$ & LeakyReLU, $\alpha=0.2$ \\
					\hline
					Training frequency & every 2 episode & every 100 steps \\
					\hline
					Optimizer & Adam & Adam \\
					\hline
					Learning rate & 0.001 & 0.000005 \\
					\hline
					$\beta$ & 1.0 & 10, 30, 100, 300 \\
					\hline
				\end{tabular}
			}
			\caption[]{Hyperparameters of GAN.}\label{tab:gan}
		\end{center}
	\end{table*}

	\begin{table*}[!htbp]
		\begin{center}
			{
				\begin{tabular}{c | c } 
					& Atari \& Mario \\
					\hline
					Grey scaling & True \\
					\hline
					Frame stack & 4 \\
					\hline
					Observation downsampling & $(84,84)$ \\
					\hline
					Action repeat & 4 \\
					\hline
					Extrinsic reward clipping & [-1,1] \\
					\hline
					Terminate when life loses & True \\
					\hline
					Skip frame & 4 \\
					\hline
				\end{tabular}
			}
			\caption[]{Preprocessing details for the environments of Atari and Mario.}\label{tab:atari}
		\end{center}
	\end{table*}

	\begin{algorithm*}[tb]
		\caption{DQN-GAEX pseudo-code}
		\label{alg:dqn_gae}
		\begin{algorithmic}[1] 
			\STATE $K_1\leftarrow$ training frequency of DQN
			\STATE $K_2\leftarrow$ training frequency of GAN compared to DQN
			\STATE $N\leftarrow$ batch size
			\STATE $\gamma\leftarrow$ discount factor
			\STATE initialize $\theta_{DQN}$, $\theta_{D}$ and $\theta_{G}$
			\FOR{episode = $1,2,..., L$}
			\STATE $t=0$
			\STATE sample initial state $s_0$
			\WHILE{$s_t$ is not the terminal state}
			\STATE sample $a_t \leftarrow \epsilon\text{-greedy}(s_t)$
			\STATE take action $a_t$ and sample $r_t^e,\ s_{t+1}$
			\STATE store $(s_t, a_t, r^e_t, s_{t+1})$ into replay buffer $\mathit{M}$
			\IF {$t\ \%\ K_1 =0$}
			\STATE sample $\left(s_{(j)}, a_{(j)}, r^e_{(j)},s'_{(j)}\right)$ uniformly from $\mathit{M}$, $j=1,...,N$
			\STATE $ r^i_{(j)} \leftarrow f\left(D_{\theta_D}\left(\phi\left(s'_{(j)}\right)\right)\right)$, $j=1,...,N$
			\STATE optimize $\theta_{DQN}$ using sampled transitions with $r_{(j)} = r^e_{(j)} + r^i_{(j)}$
			\IF {$t\ \%\ K_2=0$}
			\STATE sample $z_{(j)}\sim \mathcal{N}(0,1),\ j=1,...,N$
			\STATE update $\theta_D$ by ascending in the direction: 
			$$\quad\quad\quad\quad\quad\quad\quad\quad\quad\quad\nabla_{\theta_D}\frac{1}{N}\sum_{j=1}^{N}\left[ \log D_{\theta_D}\left(\phi\left(s'_{(j)}\right)\right) + \log\left(1-D_{\theta_D}\left(G_{\theta_G}\left(z_{(j)}\right)\right)\right) \right]$$
			\STATE update $\theta_G$ by descending in the direction:
			$$\quad\quad\quad\quad\quad\quad\quad\quad\quad\quad\nabla_{\theta_G}\frac{1}{N}\sum_{j=1}^{N}\left[ \log\left(1-D_{\theta_D}\left(G_{\theta_G}\left(z_{(j)}\right) \right) \right) \right]$$
			\ENDIF
			\ENDIF
			\STATE $t\leftarrow t+1$
			\ENDWHILE
			\ENDFOR
		\end{algorithmic}
	\end{algorithm*}
	
\end{document}